\documentclass[a4paper,fleqn]{cas-dc}
\usepackage{graphicx} % Required for inserting images
\usepackage{amssymb}  % Provides various useful mathematical symbols
\usepackage{array}
\usepackage{tabularx}
\usepackage{comment}
\usepackage{amsmath}
\usepackage{tikz}
\usepackage{subcaption}
\usepackage{pgfplots}
\usepackage{layouts}
\usetikzlibrary{positioning}
\usetikzlibrary{fit}
\usetikzlibrary{backgrounds}
\usepackage{multirow}
\usepackage{booktabs}
\usepackage{adjustbox}
\usepackage{tabularray}
\usepackage{threeparttable}
\UseTblrLibrary{booktabs}
\usepackage{lipsum}
\usepackage{xcolor}
\usepackage[square,numbers]{natbib}
\usepackage[normalem]{ulem}

\def\tsc#1{\csdef{#1}{\textsc{\lowercase{#1}}\xspace}}
\tsc{WGM}
\tsc{QE}
\makeatletter
\@fleqnfalse
\@mathmargin\@centering
\makeatother

\newcommand{\Nc}{\mathcal{N}}
\newcommand{\Vc}{\mathcal{V}}

\newcommand{\xv}{\mathbf{x}}
\newcommand{\Xv}{\mathbf{X}}
\newcommand{\zv}{\mathbf{z}}
\newcommand{\Zv}{\mathbf{Z}}

\newcommand{\cv}{\mathbf{c}}
\newcommand{\Cv}{\mathbf{C}}
\newcommand{\pred}{\hat{y}}

\newcommand{\hv}{\mathbf{h}}
\newcommand{\Hv}{\mathbf{H}}

\newcommand{\ev}{\mathbf{e}}
\newcommand{\Ev}{\mathbf{E}}

\newcommand{\SingleViewCNN}{\text{CNN}_{sv}}
\newcommand{\SingleViewBackbone}{B_{sv}}
\newcommand{\SingleViewHead}{H_{sv}}

\newcommand{\MultiViewCNN}{\text{CNN}_{mv}}
\newcommand{\MultiViewBackbone}{B_{mv}}
\newcommand{\MultiViewHead}{H_{mv}}
\newcommand{\ViewPool}{ViewPool}

\newcommand{\CentralizedInferenceTag}{CI}
\newcommand{\SelectiveCentralizedInferenceTag}{SCI}
\newcommand{\SelectiveCentralizedInferenceEmbeddingTag}{SCI-E}
\newcommand{\SelectiveCentralizedInferenceColorHistTag}{SCI-CH}
\newcommand{\EnsembleInferenceTag}{EI}
\newcommand{\SelectiveEnsembleInferenceTag}{SEI}
\newcommand{\SelectiveEnsembleInferenceEmbeddingTag}{SEI-E}
\newcommand{\SelectiveEnsembleInferenceColorHistTag}{SEI-CH}

\begin{document}
\let\WriteBookmarks\relax
\def\floatpagepagefraction{1}
\def\textpagefraction{.001}

\shorttitle{Edge-device Collaborative Computing for Multi-view Classification}    

\shortauthors{M. Palena, T. Cerquitelli, C.\,F. Chiasserini}  

\title [mode = title]{Edge-device Collaborative Computing for Multi-view Classification
}  

% \tnotemark[<tnote number>] 



\author[2]{Marco Palena}

% \cormark[<corr mark no>]

% \fnmark[<footnote mark no>]

% \ead{<email address>}

% \ead[url]{<URL>}

\author[1]{Tania Cerquitelli}
\author[1,2]{Carla Fabiana Chiasserini}

% \credit{<Credit authorship details>}

\affiliation[1]{organization={Politecnico di Torino},
            addressline={Corso Duca degli Abruzzi, 24}, 
            city={Torino},
            postcode={10129}, 
            state={TO},
            country={Italy}}

\affiliation[2]{organization={CNIT},country={Italy}}

% % Footnote of the second author

% % Email id of the second author
% \ead{}

% % URL of the second author

% % Credit authorship


\begin{abstract}
Motivated by the proliferation of Internet-of-Thing (IoT) devices and the rapid advances in the field of deep learning, there is a growing interest in pushing deep learning computations, conventionally handled by the cloud, to the edge of the network to deliver faster responses to end users, reduce bandwidth consumption to the cloud, and address privacy concerns. However, to fully realize deep learning at the edge, two main challenges still need to be addressed: (i) how to meet the high resource requirements of deep learning on resource-constrained devices, and (ii)  how to leverage the availability of multiple streams of spatially correlated data, to increase the effectiveness of deep learning and improve application-level performance. 
To address the above challenges, we explore \emph{collaborative inference} at the edge, in which edge nodes and end devices share correlated data and the inference computational burden by leveraging different ways to split computation and fuse data. Besides traditional centralized and distributed schemes for edge-end device collaborative inference, we introduce {\em selective schemes} that decrease bandwidth resource consumption by effectively reducing data redundancy.  As a reference scenario, we focus on multi-view classification in a networked system in which sensing nodes can capture overlapping fields of view. The proposed schemes are compared in terms of accuracy, computational expenditure at the nodes, communication overhead, inference latency, robustness, and noise sensitivity. Experimental results highlight
that 
%(1) different ways to split computation between nodes lead to different trade-offs between prediction accuracy and resource consumption; 
%(2) fusing features at a centralized location increases accuracy by 15-18\%, on average, with respect to fusing local predictions, but involves a higher transmission cost, and (3) that 
selective collaborative schemes can achieve different trade-offs between the above  performance metrics,  with some of them bringing substantial communication savings (from 18\% to 74\% of the transmitted data with respect to centralized inference) while still keeping the inference accuracy well above 90\%.
%We then characterize each scheme in terms of different networking and data-related properties, highlighting their benefits and the application- and system-level needs.
%they can or cannot address, as well as identifying  some practical scenario to which each of them may conform well.
%We conclude by providing some insights on the relevant challenges that still need to be addressed in order to fully unlock the potential of the proposed collaborative inference schemes. 
\end{abstract}


% \begin{highlights}
% \item 
% \item 
% \item 
% \end{highlights}

% Keywords

\maketitle

\section{Introduction}
\label{sec:intro}

% Context introduction
Accelerated by the remarkable success of Internet of Things (IoT) technologies, more and more practical environments are being equipped with sensors connected to mobile and IoT devices. Prominent examples of such environments include smart cities, smart transportation systems, and smart factories. The large amounts of data collected by the above devices, in conjunction with recent breakthroughs in deep learning, are driving the proliferation of intelligent applications and services. In addition, the edge computing paradigm is increasingly shifting computing loads from the core to the edge of the network to deliver faster responses to end users, reduce bandwidth consumption towards the cloud, and address privacy concerns.

However, several significant challenges still remain to be addressed in order to fully realize deep learning at the edge. On the one hand, increasingly complex inference tasks require highly parametrized deep neural networks (DNNs) to be executed, which crave for computational and memory resources. On the other hand, most edge devices that perform inference tasks in real-time have limited computation, memory, energy, and communication capabilities. Furthermore, IoT devices are often deployed with some degree of redundancy or overlap, resulting in data streams exhibiting significant spatial  correlation. Being able to fuse data coming from multiple sources is key to achieving better prediction accuracy in many fine-grained inference tasks, although it comes at a higher communication and orchestration cost. 
%\MP{Add some citations.}
% Such complex scenarios are often characterized by the following challenges:
% \noindent
% \textbullet\, Increasingly complex inference tasks need to be executed in real time and at edge nodes;
% \noindent
%   \textbullet\,  On the one hand, such inference tasks  involve the execution of computationally-heavy inference pipelines based on deep neural networks (DNNs), which crave for computational and memory resources.  On the other hand,  edge nodes are  resource-constrained in terms of computation, memory, energy, and communication capabilities;
%    \noindent
%    \textbullet\,  Sensors are  deployed with some degree of redundancy or overlap, resulting in streams of data exhibiting significant spatio-temporal correlation. 
%     %For instance, multiple cameras may have partially overlapping fields of view (FoV), multiple arrays of microphones may capture the same sound with some time delay, etc.
% \noindent  
Considering these trends, it is critical to envision efficient mechanisms that allow for the \emph{collaborative} execution of increasingly complex inference tasks at the edge of the network, with the different edge nodes sharing data as well as the computational burden.  

% Problem description
In this paper, we take the above challenge, focusing on the relevant case of computer vision tasks applied to images collected by different end devices. 
Computer vision tasks, enabling machines to derive meaningful information from digital images, have been at the forefront of deep learning applications due to their important role in solving problems arising from various fields.
Differently from existing works tackling the orchestration of DNNs at the edge \cite{malandrinoMatchingDNN2022, chenDeepLearningEdge2019} or the offloading of such tasks from end devices to the edge \cite{chenDeepLearningEdge2019,pulighedduSemanticORAN2023}, our first goal is to \emph{leverage the redundancy or overlap of the images} collected by such cameras to increase the accuracy of the inference while reducing the computational and communication load that such inference implies. 
Our second goal is to \emph{investigate the benefits and the hurdles of different collaborative approaches} between edge servers and end devices capturing the images, each reflecting a different level of cooperation and split of the computational/communication burden at the edge and at the devices.
Collaboration can involve sharing features or inferences and/or splitting computation (i.e., distributing the machine learning operations) between the different nodes. Thus, it is crucial to assess their performance from the point of view of both inference quality and consumption of radio/computational resources. 

Specifically, we envision a networked system with per-node sensing and/or processing capabilities, in which sensing nodes are equipped with cameras with partially overlapping fields of view. At distinct time instants, the sensing nodes collect a set of spatially  correlated images (views) pertaining the same object observed from distinct viewpoints. A real-world example of such a scenario is that of a smart city in which cameras embedded in the road infrastructure collect images over the same stretch of road, and cooperate with an edge server to solve a computer vision task based on the collected data. 
For concreteness, in our investigation, we focus on the specific task of multi-view classification~\cite{sunSurveyMultiviewMachine2013,seelandMultiviewClassificationConvolutional2021}, and compare different collaborative approaches using a state-of-the-art DNN architecture~\cite{suMultiviewConvolutionalNeural2015} for the problem. Nevertheless, we advocate that the approaches we analyse are both task- and architecture-agnostic and, thus, can be easily adapted to different inference pipelines and scenarios.

% Contributions
Our main contributions are, therefore, as follows:

\noindent
  \textbullet\, We provide a formalization of the networked image-capturing system and a model for describing different approaches to collaborative inference tasks in such a scenario;
    
     \noindent
  \textbullet\, We consider and formally define different schemes for edge-end devices collaborative inference, specifying which building blocks compose the inference processing framework and where they are deployed. Importantly, besides traditional centralized and distributed schemes, we propose novel {\em selective} schemes that realize different data fusion and computational split strategies. 
In these schemes, each sensing node  locally decides whether or not its view should contribute to the inference task at hand based on some contextual information, thus restricting inference to a subset of the most relevant views. 
  By doing so, our  schemes  enable   different trade-offs between prediction quality and consumption of computational and bandwidth resources;

     \noindent
  \textbullet\, We experimentally analyze the different schemes and compare them in terms of accuracy, computational expenditure at the nodes, communication overhead, inference latency as well as robustness, and noise sensitivity. Our results highlight %nteresting  trade-offs between  accuracy and resource utilization that characterize the different schemes, indicating 
  that (1) not all views are always necessary to achieve high-quality predictions,  instead, substantial communication savings (from 18\% to 74\% of the transmitted data) can be obtained discarding some views while still keeping the accuracy well above 90\%; (2) even when no views are discarded, allowing inference to be partially  computed at the end devices still yields an average accuracy ranging between  71.92\% and 83.75\%, in average, while 
  reducing bandwidth consumption by one or more orders of magnitude with respect to centralized inference; 
  
    \textbullet\, We characterize each scheme in terms of different networking and data-related properties, highlighting their benefits along with the application- and system-level needs they can or cannot address. Furthermore, we discuss possible practical application scenarios for each scheme.

% Outline
The paper is organized as follows. Sec.~\ref{sec:related} discusses some relevant related works. Sec.~\ref{sec:prelim} introduces the networked system we consider and formalizes the problem we address. Sec.~\ref{sec:schemes} describes the different collaborative inference schemes we investigate. An experimental comparison of the proposed schemes focused on relevant metrics is presented in Sec.~\ref{sec:exp}. Finally, Sec.~\ref{sec:discussion}  provides an additional qualitative comparison between the proposed schemes, discusses suitable practical applications for each of them, and concludes the paper envisioning future research directions.

\section{Related Work}
\label{sec:related}
% Works on collaborative inference at the edge

A large body of works have focused on supporting AI tasks at the edge, a research field known as \emph{edge intelligence}~\cite{renSurveyCollaborativeDNN2023,zhou2019}. A common solution to this problem is computation offloading, in which end devices offload the inference computational burden (or part of it) to another entity, being it a cloud server, an edge server, or another end device. These approaches are often based on model partitioning, splitting a multi-layered DNN across the different entities participating in the execution of the inference task.  Existing approaches can be broadly categorized into (i) strategies that offload computation to a single centralized entity (either on the cloud~\cite{eshratifar2018,kang2017} or at the edge~\cite{wangRFSensing2018,chenDeepLearningEdge2019}), and (ii) strategies that partition the DNN model across multiple end devices, enabling joint computation either in a purely decentralized~\cite{malkaDecentralized2022,zeng2020} or centrally-orchestrated way~\cite{maoSurveyOnMobile2017}. 
In both cases, each participating device retains only a portion of the DNN layers and transmits its output features, potentially compressed to minimize overhead~\cite{cohenLightweightCompression2021}, to another node. The primary drawback of this method is that no single device can perform inference independently; the entire group of devices sharing the DNN layers must be available.
To overcome these limitations, some studies have focused on enabling inference at edge devices via DNN compression techniques such as parameter pruning, parameter quantization, and knowledge distillation~\cite{choudharyComprehensiveSurveyModelCompression2020}. Importantly, such techniques, although orthogonal to the collaborative inference schemes we investigate, can be integrated with those we consider to further reduce computation and communication overhead.

It is worth it to remark, however, that most of the above existing studies focus on inference tasks arising from a single end device, thus considering other nodes merely for their additional computational resources. Conversely, in our work, we consider a scenario in which the inference task can make use of the information generated by  multiple end devices, each located at a different position. In such a scenario, collaboration among different nodes entails not only  sharing the computational burden but also  fusing  correlated data. The need, and opportunity, to leverage spatially correlated data from different sources raises additional challenges such as deciding whether or not all captured data is actually needed to perform the inference, as the consumption of computational and network resources should be minimized. Although some recent work~\cite{hao2023, wangMultiAgentSystemsCollaborative2024} has considered a multi-agent collaborative inference scenario, where a single edge server coordinates the inference of multiple UEs,   inference based on spatially correlated data captured by different nodes has not been addressed. Rather, the goal of the studies in \cite{hao2023, wangMultiAgentSystemsCollaborative2024}  is to  minimize inference latency when multiple end devices offload their inference computation to the same edge server.

%Works on fusion of spatio-temporally correlated data
On the other hand, the works that in the context of IoT focus on strategies to fuse data from multiple sources  leveraging on  spatio-temporal correlations~\cite{yaoDeepSense2017,qiuKestrelVideoAnalytics2018,yanDeepMultiviewLearning2021}  consider that the inference is performed at a centralized location and are not geared toward balancing the usage of local computational resources and bandwidth. 
As for state-of-the-art studies on multi-view classification, these  
are tailored to optimizing deep learning multi-view models to achieve higher prediction accuracy. For instance, in~\cite{silvaMultiViewFineGrained2021} Silva \emph{et al.} train an auxiliary classifier to optimize the feature extractors of each view and apply the modified multi-view architecture to the task of automatic toll collection. In~\cite{seelandMultiviewClassificationConvolutional2021}, 
 Seeland and M\"{a}der explore different view fusion strategies. Differently from these works, our focus is on efficiently partitioning an existing multi-view architecture across different nodes rather than optimizing the individual stages of these architectures.

Finally, we remark that there exists an extensive body of research on techniques exploiting the temporal correlation between views, such as frame differentiation~\cite{chenGLIMPSEContinuousRealTime2016}. This is  an orthogonal research direction with respect to our work, which,   besides tackling computational split strategies, focuses instead  on  exploiting the spatial correlation between views captured by nodes at different positions so as to reduce information redundancy and save bandwidth.

\section{System Model and Inference Task}
\label{sec:prelim}

In this section we describe the network and computing system under study (Sec.\,\ref{sub:system}). Then, we provide some background on single-view and multi-view classification, as well as on the architecture of the deep learning models used to perform them (Sec.\,\ref{sub:problem}). Finally, we briefly introduce color histograms, a lightweight image descriptor that will be used in some of the proposed schemes  (Sec.\,\ref{sub:color_hist}). The notation we use throughout the paper is summarized in Table~\ref{tab:notation}.

\begingroup
\setlength{\tabcolsep}{5pt}
\renewcommand{\arraystretch}{1.25}
\begin{table}
\centering
\caption{Summary of notation}
\begin{tabularx}{\linewidth}{l l}
 {\bf Definition} &  {\bf Meaning} \\
 \hline
 %$\Gc = (\Nc, \Lc)$ & Network graph\\
% $\Lc \subseteq \Nc \times \Nc$ & Set of communication links\\
 $\Vc = \{n_1, \dots, n_V \}$ & Set of source nodes\\
 $n^* \in \Nc$ & Central controller node\\
  $\Nc = \Vc \cup \{n^*\}$ & Set of physical nodes\\
% $n_i \in \Vc$ & $i$-th source node\\
 \hline
% $\Xv = \InputSpace$ & Input (image) space\\
 $y \in Y {=} \{y_1, \dots, y_K\}$ & Class label \\
 $\xv_i(t) \in \Xv$ & View captured by node $n_i$ at time $t$ \\
 %$\xv(t) = [\xv_i(t), \ldots, \xv_V(t)]$ & Collection of views captured at time $t$\\
 %$\zv(t) \in \Zv$ & Intermediate representation, with $\zv_i(t)$ corresponding to  $\xv_i(t)$\\
 $\zv_i^{(j)}(t) \in \Zv$ & $j$-th intermediate representation of $\xv_i(t)$\\
 %\zv(t) = [\zv_i, \ldots, \zv_{V'}]$ & Collection of intermediate representations\\
% $\Cv$ & Context space\\
 $\cv(t) \in \Cv$ & Context   at time $t$ \\
% $\mv_i^*(t)$ & Message sent from $n^*$ to $n_i$ at time $t$\\
% $\mv_i(t)$ & Message sent from $n_i$ to $n^*$ at time $t$\\
 $\pred(t) \in Y$ & Prediction made at time $t$\\
 \hline
% $\Ev$ & Embedding space\\
 $\ev \in \Ev$ & Embedding of an image $\xv$\\
% $\ev_{mv} \in \Ev$ & Multi-view embedding\\
 $\xv_{V} = [\xv_1, \ldots, \xv_V]$ & Collection of $V$ views\\
 $\ev_{V} = [\ev_1, \ldots, \ev_V]$ & Collection of $V$ embeddings\\
 $\SingleViewCNN: \Xv \to \pred$ & Single-view CNN model\\
 $\SingleViewBackbone: \Xv \to \Ev$ & Single-view feature extraction network\\
 $\SingleViewHead: \Ev \to \pred$ & Single-view classification network\\
 $\MultiViewCNN: \Xv^V \to \pred$ & Multi-view CNN model\\
 $\MultiViewBackbone: \Xv^V \to \Ev$ & Multi-view feature extraction network\\
 $\MultiViewHead: \Ev \to \pred$ & Multi-view classification network\\
 $\ViewPool$ & View pooling function\\
 \hline
% $\Hv = \mathbb{R}^{B \times B}$ & Color histogram space with $B$ bins\\
 $\hv_i(t) \in \Hv$ & Color histogram of the view $\xv_i(t)$\\
 $hist: \Xv \to \Hv$ & Color histogram function\\
 \hline
\end{tabularx}
\label{tab:notation}
\end{table}
\endgroup

\begin{figure}
    \centering
    \resizebox{!}{40mm}{
        \includegraphics{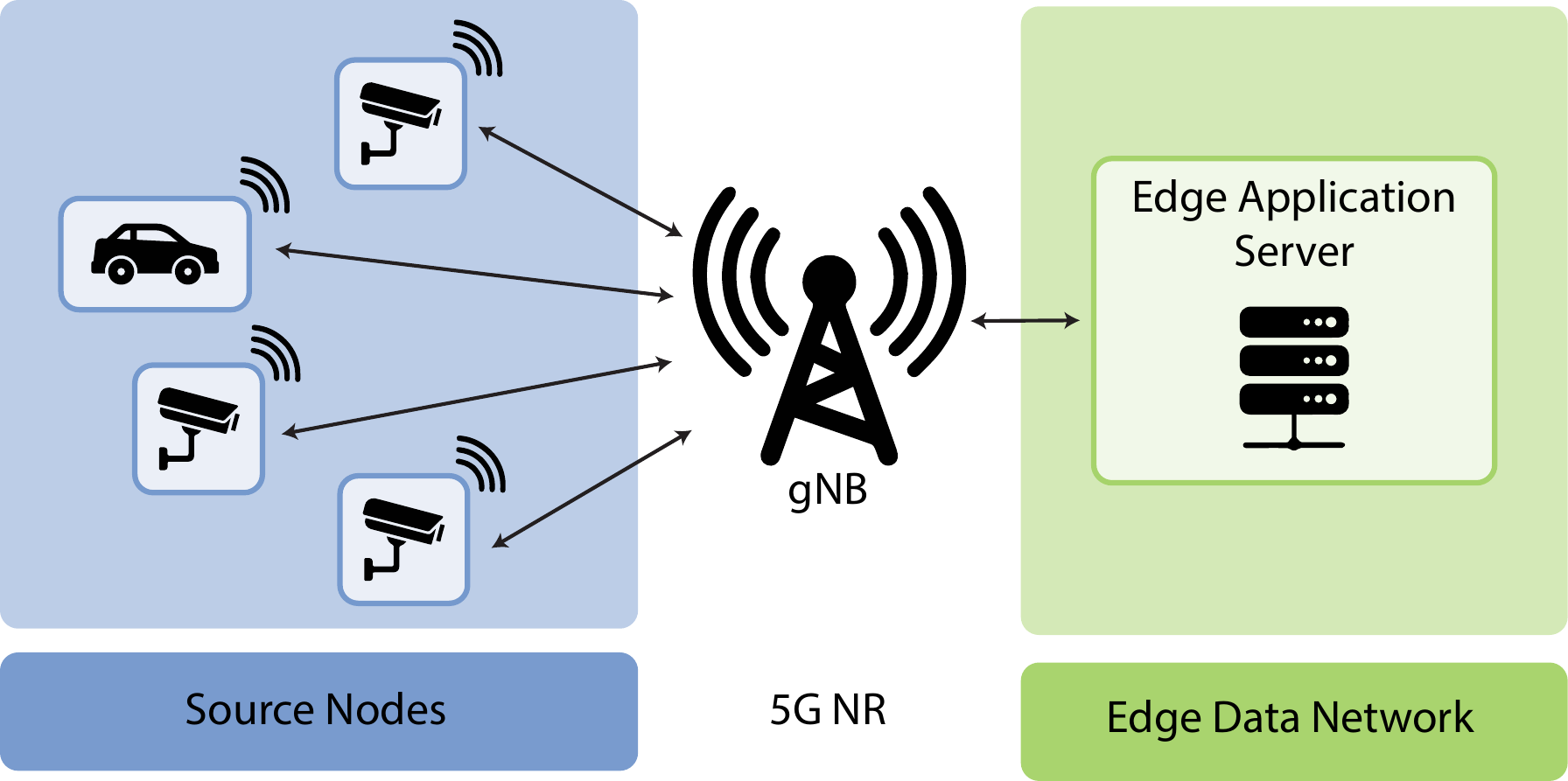}
    }
    \caption{Example of our reference system scenario in the 5G ecosystem.} 
    \label{fig:system_model}
\end{figure}

\subsection{Network and computing system\label{sub:system}}

%We The system we envision is modelled as a directed graph \mbox{$\mathcal{G} {=} (\mathcal{N}, \mathcal{L})$}, with $\mathcal{N}$ and $\mathcal{L}$ representing, respectively, the \emph{physical nodes} and the \emph{communication links} connecting the nodes. The set of nodes $\Nc$ is composed of a set $\Vc{=}\{n_1, \dots, n_V \}$ of \emph{source nodes}, capturing images, and a \emph{central controller} node $n^*$. 

We consider $\Vc{=}\{n_1, \dots, n_V \}$, a set of \emph{source nodes} composed of $V$ edge devices equipped with a camera, a radio interface, as well as computational and memory resources.  
 Source nodes may capture images of the same object, which differ in the perspective with which the object is captured, or in quality, size, or resolution. 
We refer to such an image,  displaying an object captured by a source node, as a \emph{view}. 
 Each source node is also connected, for instance through a 5G link, with an edge server hosting a \emph{central controller}, with the latter being denoted by $n^\star$. We indicate  the set of all system nodes with
 $\Nc$. Both source nodes and central controller can collaborate to execute multi-view classification tasks on the images captured by the source nodes.  
Time is assumed to be discretized, with each discrete time period $t$ corresponding to the time interval during which the source nodes capture their images, and the system has to execute the classification task.  
%{\color{purple} In the following, we denote with $m_i^*(t)$ the messages sent at time $t$ by the central controller to the generic source node $n_i$, and with $m_i(t)$ the messages sent from a source node $n_i$ to the central controller. 
%Whenever clear from the context, we will omit the time indication $t$ to streamline the notation. }}

Importantly, the central controller acts as a data aggregator and coordinator for the collaborative inference task.
Once the result of the multi-view classification task is obtained, we consider that the central controller returns the prediction to the source nodes, which feed it as input to their local applications. This is required whenever source nodes are autonomous mobile nodes, like connected vehicles and unmanned aerial vehicles, which   need the classification result as input to their local application for  navigation purposes.  Fig.~\ref{fig:system_model} provides an example of the considered system in the 5G ecosystem, where source nodes may be connected vehicles as well as road infrastructure cameras communicating with an edge application server via 5G.
Further details on the main system components are given  below.  

\begin{figure}
    \centering
    \resizebox{\columnwidth}{!}{
        \includegraphics{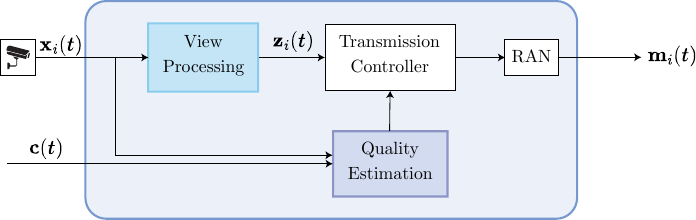}
    }
    \caption{Schematic representation of  source node $n_i$.}
    \label{fig:source_node}
\end{figure}

\paragraph{\textbf{Source nodes.}}
At time $t$, each source node $n_i$ collects a \emph{view} $\xv_i(t)$ and receives the current \emph{context} $\cv(t)$ from the central controller. The view is processed locally to obtain one or more intermediate representations $\zv_i(t)$, which, the node can decide to send to the central controller or discard based on the contextual information received (see Sec.\,\ref{sec:schemes}). When multiple intermediate representations are computed for a given view, we indicate  with  $z_i^{(j)}(t)$  the $j$-th representation.  Fig.~\ref{fig:source_node} provides the schematic representation of the generic source node $n_i$, which includes  the following modules:

\textbullet\, \emph{View Processing}, responsible for applying any local transformation on the captured view $\xv_i(t)$ and for generating the intermediate representation $\zv_i(t)$. The type of processing it performs depends on the particular inference scheme, and it can range from no processing, to feature extraction, to full single-view classification. 
%\TC{Qui bisognerebbe specificare meglio che tipo di elaborazione viene fatta.} \MP{Aggiunta una descrizione del processing.}

\textbullet\,
 \emph{Quality Estimation}, which takes as input the current view and context and determines whether or not the intermediate representation of the view is worth being transmitted to the central controller. We define a view's \emph{quality} as a measure of similarity between the view and the current context, representing the collective set of features captured by the network at the previous or current time instant. Internally, the quality estimation module thus computes a similarity measure and tests it against a threshold. %\TC{Qui bisognerebbe definire che cosa intendiamo per qualità.} \MP{Aggiunta una descrizione della qualità.}

\textbullet\,  \emph{Transmission Controller},  handling the communication with the central controller via the radio interface. 
%\TC{Qui bisognerebbe capire se è un blocco intelligente, perche applica qualche logica, oppure no.} \MP{Aggiunto un commento in fondo alla sezione.}

%A \emph{Context Manager} module may provide contextual information useful to determine whether or not the current (processed) view is worth being transmitted to the central controller. The \emph{context} is derived based on information received from the central controller and is described by a tensor 

% Upon collecting a new view, the node applies a view processing function $f_\Vc: \Xv \to \Ev$ to it, obtaining an intermediate view representation $\ev_i(t) {\in} \Ev$, with $\Ev$ being an \emph{intermediate view space}.

\begin{figure}
    \centering
    \resizebox{\columnwidth}{!}{
        \includegraphics{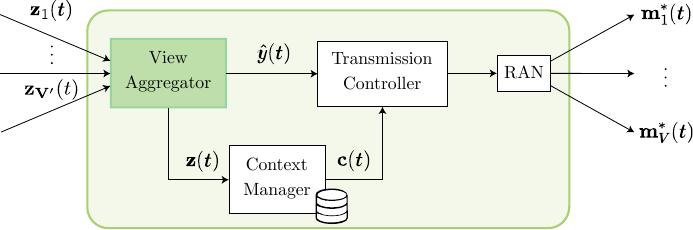}
    }
    \caption{Schematic representation of the central controller  $n^\star$.}
    \label{fig:controller_node}
\end{figure}

% Detailed formal description of the central controller node
\paragraph{\textbf{Central controller.}}
At time $t$, the central controller $n^\star$ receives  $V' {\leq} V$ intermediate view representations, each from a distinct source node $n_i$ and represented by the corresponding tensor $\zv_i(t)$. Such tensors are aggregated into a higher dimensional tensor $\zv(t)$, which is then processed to obtain the current context $\cv(t)$ as well as the final prediction $\pred(t)$. The latter is then sent back to the source nodes. Fig.~\ref{fig:controller_node} provides a schematic representation of the controller node, comprising the following additional modules:

\textbullet\, \emph{View Aggregator}, which aggregates the intermediate representations received from the source nodes and derives the final prediction;

\textbullet\, \emph{Context Manager},  determining the current context $\cv(t)$ based on the intermediate view representations received from the source nodes at the current or previous time instants;

% \textbullet\,  \emph{Transmission Controller} handling communication with the source nodes via the radio interface.

The modules introduced above for the source nodes and the central controller will then be further specified for each of the collaborative inference schemes we analyze in Sec.~\ref{sec:schemes}. \emph{Transmission Controller} modules are assumed not to include any inference scheme-related logic and, therefore are implemented similarly for each scheme.

\subsection{Single- and multi-view inference tasks\label{sub:problem}}

Below, we define our reference inference tasks, namely, single-view and multi-view image classification.
For each of them, we also present the high-level architecture of state-of-the-art deep learning pipelines that can be used to execute the task.

\begin{figure}[tbh]
    \centering
    \resizebox{\columnwidth}{!}{
        \includegraphics{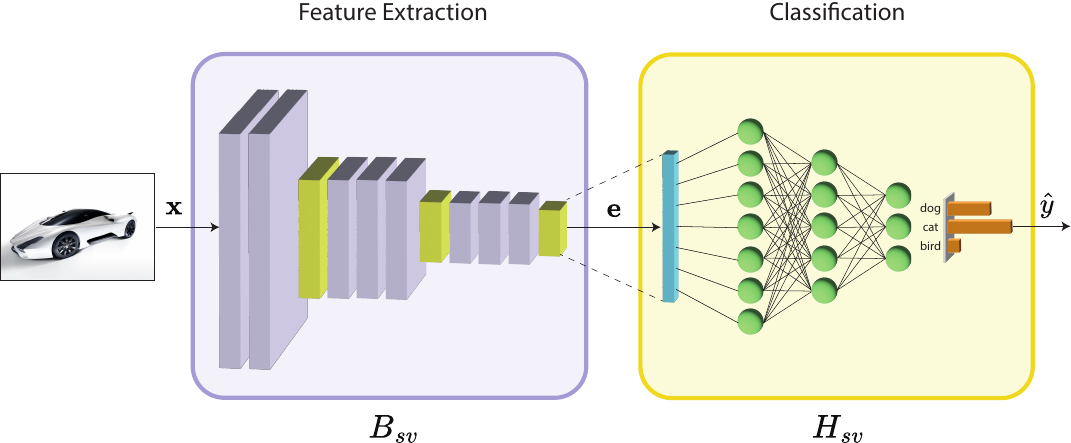}
    }
    \caption{Architecture of a deep single-view CNN.}
    \label{fig:svcnn}
\end{figure}
\begin{figure}[tbh]
    \centering
    \resizebox{\columnwidth}{!}{
        \includegraphics{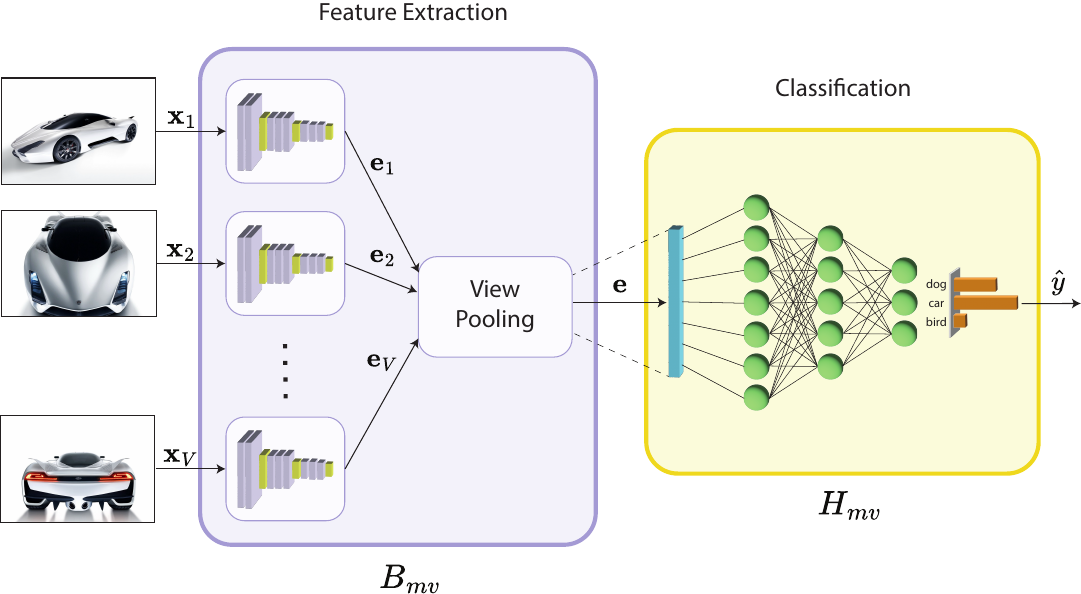}
    }
    \caption{Architecture of a deep multi-view CNN.}
    \label{fig:mvcnn}
\end{figure}

\paragraph{\textbf{Single-view image classification.}}
\label{par:image_classification}
% Formalization of the single-view image classification task
Given an input image $\xv{\in}\Xv$ and a set of K labels $Y {=} \{y_1, \ldots, y_K\}$ representing classes of interest, the single-view image classification task consists in inferring which label $y {\in} Y$ to assign to $\xv$, i.e., a single-view image classifier is a function mapping input images to class labels.  
State-of-the-art image classifiers are often implemented through complex, resource-craving convolutional neural networks (CNNs)~\cite{lecun2015deep, krizhevskyImagenetClassification2012}.
A CNN for single-view image classification $\SingleViewCNN$ is a model composed of two parts: a convolutional \emph{feature extraction network} (backbone) and a fully-connected \emph{classification network} (head), as shown in Fig.~\ref{fig:svcnn}. The feature extraction network, denoted by $\SingleViewBackbone$, consists of a sequence of 2D convolutional blocks and pooling layers and is responsible for extracting and encoding visual features from the input image into an \emph{image embedding} $\ev {\in} \Ev$. The \emph{embedding space} $\Ev$ is a lower-dimensional space than $\Xv$, making the image embedding a more compact representation of the visual features of the image. The classification network, denoted by $\SingleViewHead$, is composed of several fully connected layers followed by a softmax activation function, mapping the image embedding $\ev$ to a class label $\pred$.
%The \emph{input space} $\Xv$ is the space of 3-dimensional tensors $\mathbb{R}^{\mathit{W} {\times} \mathit{H} {\times} \mathit{C}}$ with $C$ being the number of channels and $W {\times} H$ being the \emph{pixel resolution} of the input images. 
% Description of the single-view CNN architecture

\paragraph{\textbf{Multi-view image classification.}}
\label{par:multi_view_classification}
% Formalization of the multi-view image classification task
This task can be performed when multiple input images describing the same object from distinct perspectives are available.  A set of views of the same object is called a \emph{multi-view collection} (or simply a collection). 
Multi-view image classification extends of the image classification task to a multi-view collection $\xv_{V} {=} [\xv_1, \ldots, \xv_V ] {\in} \Xv^{V}$.
%Given a collection of $V$ input images  and a set of $K$ classes represented by labels $Y$, the multi-view classification task consists in inferring which label $y{\in}Y$ to assign to $\xv_{V}$, i.e., the classifier is a function mapping input collections to class labels. 
%The input space $\Xv^{V}$ can be seen as the 4-dimensional tensor space $\mathbb{R}^{\mathit{V} \times \mathit{W} \times \mathit{H} \times \mathit{C}}$.
% Description of the multi-view CNN architecture
We consider deep multi-view classification models based on the MVCNN architecture~\cite{suMultiviewConvolutionalNeural2015}, which uses a single-view image classification CNN as a backbone (see Fig.~\ref{fig:mvcnn}). 
Similarly to the single-view case, a CNN for multi-view image classification $\MultiViewCNN$ consist of a convolutional feature extraction and a fully-connected classification part. The former, denoted by $\MultiViewBackbone$, takes as input a  collection of $V$ input images $\xv_{V}$ and produces as output a single \emph{multi-view embedding} $\ev_{mv} {\in} \Ev$, acting as an intermediate representation of the whole multi-view collection. This is done by applying a single-view feature extractor $\SingleViewBackbone$ to each individual view $\xv_i$ to obtain a collection of embeddings $\ev_{V} {=} [\ev_1, \ldots, \ev_V] \in \Ev^{V}$. Then, orderless aggregation of these embeddings is performed via a \emph{view pooling} function $\ViewPool$, producing a single multi-view embedding $\ev_{mv}$.
The latter  part of the network, denoted with  $\MultiViewHead$, is thus a single-view image classifier $\SingleViewHead$ that maps the multi-view embedding $\ev_{mv}$ to a class label $\pred$. It follows that a multi-view CNN model, $\MultiViewCNN$, can be obtained from a single-view one, $\SingleViewCNN$, by duplicating the feature extraction network $\SingleViewBackbone$ once for each view, adding a view pooling layer, and reusing the same classification network $\SingleViewHead$. 

\begin{figure*}
\centering
\includegraphics[width=\textwidth]{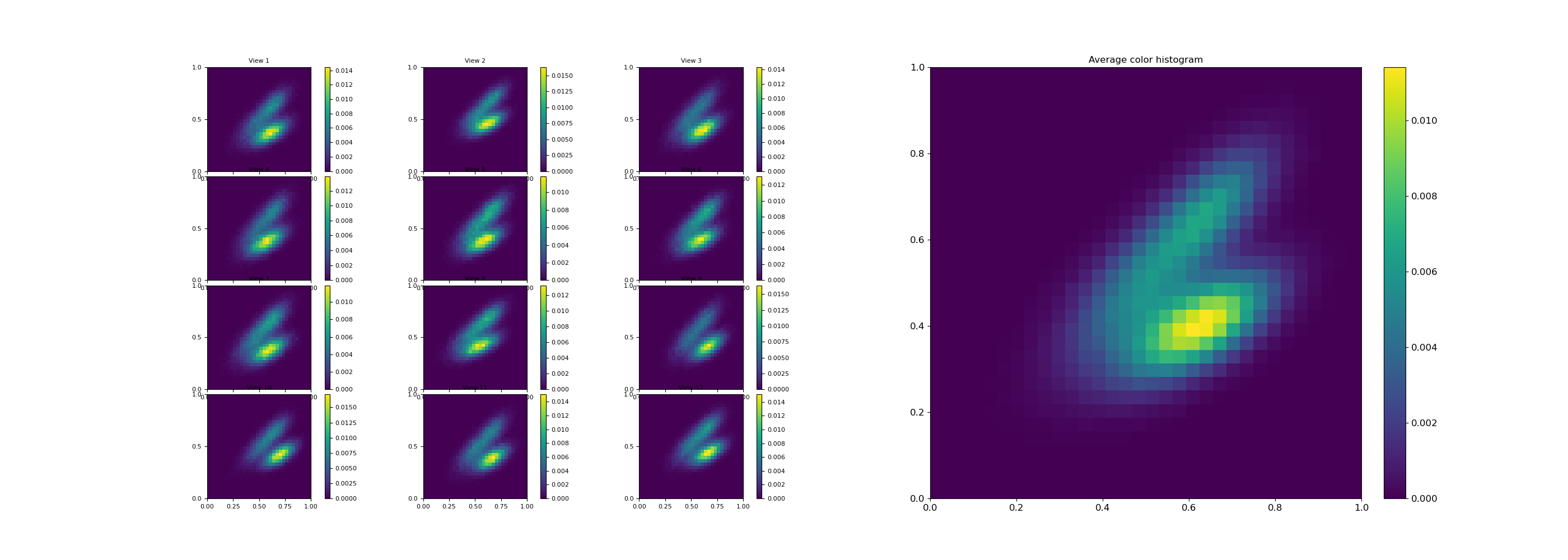}
\caption{Color histograms with 32 bins computed on 12 views of the same object (left) and their corresponding average (right).}\label{fig:color_hist}
\end{figure*}

\subsection{Color histograms}
\label{sub:color_hist}
In some of the inference schemes we describe in Sec.~\ref{sec:schemes}, we will make use of color histograms as lightweight descriptors of views. Given a view $\xv_i(t)$ and a number of bins $B$, we denote with $hist$ the function responsible for computing the color histogram $\hv_i(t)$ of $\xv_i(t)$. Color histograms are represented as 2-dimensional tensors in the space $\Hv {=} \mathbb{R}^{B \times B}$:
$\hv_i(t) = hist(\xv_i(t); B)$.  
The color histogram of a view is computed as follows:
\begin{enumerate}
\item First, the view image is converted from the RGB to the L*a*b* color space~\cite{standard2007colorimetry}, designed to be more perceptually uniform than the RGB.
\item Then the ranges of the L*a*b* color channels (a* and b*) are divided into $B$ bins   and each pixel is counted into one of the resulting $B {\times} B$ buckets. 
\item Finally, the pixel count of each bucket is normalized by the number of pixels. 
\end{enumerate} 
Fig.~\ref{fig:color_hist} provides a visual representation of the color histograms computed for a sample collection of views, as well as their average.

\section{Collaborative Inference Schemes}
\label{sec:schemes}
%Upon capturing an image, each source node has to perform an image classification inference. 
A straightforward approach to handle classification tasks in which multiple nodes sense spatially correlated data would be to have each source node perform its inference \emph{in isolation}, relying only on locally captured sensory data. This approach, however, has several drawbacks:

%\begin{itemize}
\textbullet \, Source nodes are often IoT devices with limited hardware resources and may be unable to execute entire deep learning pipelines;

\textbullet \, 
 For fine-grained image classification tasks  characterized by high inter-class similarity and a high intra-class variability (e.g., vehicle identification~\cite{yangLargeScaleCarDataset2015, silvaMultiViewFineGrained2021}), the visual information conveyed by a single image may be insufficient to make accurate predictions.
%k\end{itemize}

We aim at mitigating such drawbacks by leveraging the spatial correlation among views taken by different source nodes, as well as the collaboration between source nodes and the central controller at the edge. We, therefore, transform multiple disjoint instances of the single-view classification problem into a single instance of multi-view classification.  

To address the multi-view classification task collaboratively, the system nodes have to share data, or processing, or both. Processing may be shared by splitting and allocating the different layers of the multi-view classification network to different nodes. Depending on how the model is split, nodes will need to exchange different types of data (e.g., raw images, embeddings, class predictions, etc.), resulting in different bandwidth utilization and latency. Thus, while devising how to leverage collaboration between nodes, it is critical to jointly take into account networking and data-related aspects such as the communication capability of nodes, the reliability of the communication links, the presence of noise affecting the transmitted data, and the privacy requirements of end users. 

Below, we identify various schemes for collaborative inference based on how the processing is split among nodes and which data they share. We consider both traditional centralized and distributed schemes and introduce {\em selective} schemes that can substantially decrease network resource consumption by exploiting context information to reduce data redundancy. 
Indeed, as observed in~\cite{suMultiviewConvolutionalNeural2015}, given a multi-view collection, not all of  the views it contains may be needed to produce an accurate prediction. Some views may be particularly informative, capturing uniquely distinctive features of an object, while others may be redundant with respect to the rest of the collection and, thus, lead to unnecessary communication overhead.  
%The MVCNN model learns how to combine the views, using the more informative ones while discarding the others, with negligible impact on the prediction accuracy. However, in the distributed scenario we consider, the transmission of these redundant views may  lead to unnecessary communication overhead. 

% {
%REPETITION
% As each scheme exhibits a different trade-off among prediction accuracy, computation burden, and communication overhead,  we carefully analyze and compare the above schemes,  from a quantitative and qualitative point of view in the following sections.
% }
%We are interested in comparing the trade-offs of different collaborative inference schemes both from a qualitative (Sec.~\ref{sec:discussion}) and quantitative (Sec.~\ref{sec:exp}) point of view.  
We classify the  schemes under study into two main categories: 

%\begin{itemize}
\textbullet\, \emph{Centralized inference schemes}, in which inference takes place at the central controller based on aggregated multi-view input data obtained from the source nodes;

\textbullet\,  \emph{Ensemble inference schemes}, in which source nodes performs local inferences on their single-view input data and then local predictions are aggregated by the central controller.
For each category, we further distinguish   inference schemes into \emph{selective} and \emph{non-selective}, depending on whether source nodes leverage only the most informative views to be used for classification, or if, instead, all views are processed for executing the task. 

\subsection{Centralized inference schemes}
\label{sub:centralized}

In centralized inference schemes,
the central controller is responsible for aggregating single-view data received from the source nodes and then performing inference on the aggregated multi-view data. Source nodes, on the other hand, collect input images and may perform feature extraction. Below, we specify how the central controller and the source nodes split the computational burden and collaborate towards the multi-view image classification, when all views get processed as well as when source nodes identify and discard redundant information before views are further processed.

\begin{figure*}[t]
    \centering
    \begin{subfigure}[c]{0.41\textwidth}
        \centering
        \includegraphics[width=7cm]{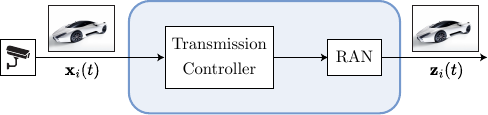}
        \caption{\label{fig:ci_scheme_source_nodes}}
    \end{subfigure}
    \begin{subfigure}[c]{0.57\textwidth}
        \centering
        \includegraphics[width=9.6cm]{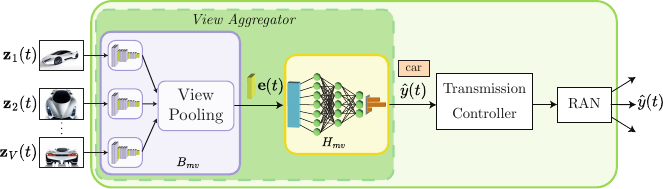}
        \caption{\label{fig:ci_scheme_controller}}
    \end{subfigure}
    \caption{Schematic representation of the operations performed by a source node (a) and the central controller (b) in the non-selective centralized inference scheme (\CentralizedInferenceTag{}). %Source nodes do not perform any processing. The central controller executes a multi-view CNN composed of its feature extraction $\MultiViewBackbone$ and classification $\MultiViewHead$ parts.
    }
    \label{fig:ci_scheme}
\end{figure*}

\subsubsection{Non-selective centralized inference}
\label{sub:ci}

In the \emph{non-selective centralized inference scheme} (\CentralizedInferenceTag), source nodes act merely as data sources, transmitting their views to the central controller.
% {
% %REPETITION
% which will then aggregate them and execute the multi-view classification model to obtain a class prediction.
% }
Specifically, the \CentralizedInferenceTag{} scheme, depicted in Fig.~\ref{fig:ci_scheme}, operates as follows:
%\begin{itemize}\item 

\textbullet\, Source nodes do not perform any processing, therefore they do not include  the \emph{View Processing} module  and the input view $\xv_i(t)$ is forwarded directly to the central controller, i.e., 
$\zv_i(t) {=} \xv_i(t)$;

\textbullet\, The central controller is responsible for performing multi-view classification; accordingly,   its \emph{View Aggregation} module  aggregates the received views into a collection and then executes a multi-view CNN  to obtain a prediction $\pred(t)$. The class prediction will then be delivered back to the source nodes.

\textbullet\, As the \CentralizedInferenceTag{} scheme is non-selective, there is no sharing of contextual information between the central controller and source nodes. Consequently, neither the \emph{Quality Estimation} module at the source nodes nor the \emph{Context Manager} module at the central controller are implemented.

\subsubsection{Selective centralized inference}
\label{subsub:sci}
In \emph{selective centralized inference schemes} (\SelectiveCentralizedInferenceTag), each source node decides whether or not to transmit its view to the central controller based on a measure of similarity between the view and the current context. The context is received from the central controller and is used as a low-dimensional representation of the set of features collectively captured by the network at the previous or current time instant. Comparing the current view to the context enables each source node to estimate how informative its view is with respect to the data sensed by the rest of the network. Only those views yielding sufficiently distinctive information with respect to the context will be forwarded to the central controller and will then contribute to the inference task, while the others   will be discarded. The central controller is still responsible for performing multi-view classification but its input will be restricted to  a subset of the most informative views. Further, we consider two variants of the \SelectiveCentralizedInferenceTag{} scheme, which suit two different types of use cases: one based on \emph{view embeddings}  and the other based on \emph{color histograms}, as detailed below.

\begin{figure*}[t]
    \centering
    \begin{subfigure}[c]{0.49\textwidth}
        \centering
        \includegraphics[width=8.3cm]{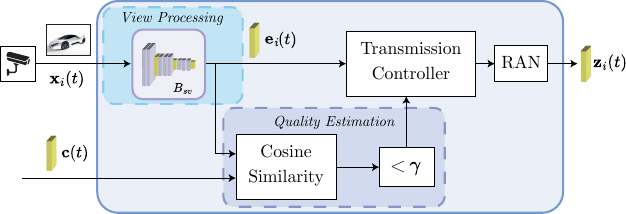}
        \caption{\label{fig:sci-e_scheme_source_nodes}}
    \end{subfigure}
    \begin{subfigure}[c]{0.49\textwidth}
        \centering
        \includegraphics[width=8.3cm]{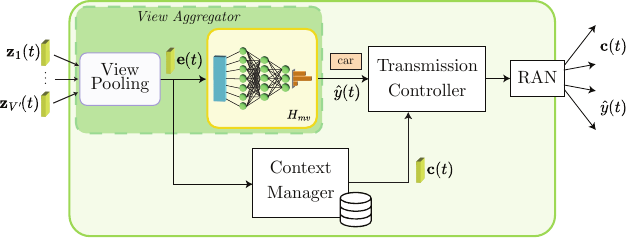}
        \caption{\label{fig:sci-e_scheme_controller}}
    \end{subfigure}
    \caption{Schematic representation of the operations performed by a source node (a) and the central controller (b) in the selective centralized inference scheme based on embeddings (\SelectiveCentralizedInferenceEmbeddingTag{}). %Source nodes execute a single-view feature extractor $\SingleViewBackbone$. The central controller pools the views and executes a fully-connected multi-view classifier $\MultiViewHead$.
    }
    \label{fig:sci-e_scheme}
\end{figure*}

\paragraph{\textbf{\SelectiveCentralizedInferenceTag{} based on embeddings.}}\label{par:sci-e}
In this case, referred to as \SelectiveCentralizedInferenceEmbeddingTag, each source node decides whether or not to send its view to the central controller by comparing it to a representation of the features collectively captured by the network at the \emph{previous} time period. Specifically, at every inference time $t$, the central controller sends to the source nodes   the pooled view embedding computed  at time $t-1$ as contextual information to discriminate views. As this scheme allows  detecting the correlation among views taken at different time (temporal change of context), it  processes the current views only if sufficiently different from what previously captured by the source nodes. 
% REPETITION
%Each source node estimates how similar the embedding of its current, locally-captured view is to the received pooled embedding  by computing the cosine between the two tensors. The result is then compared to a predefined \emph{similarity threshold} -- a configurable system parameter whose impact is analysed in Sec.~\ref{sub:threshold_eval}.   
The \SelectiveCentralizedInferenceEmbeddingTag{} scheme, depicted in Fig.~\ref{fig:sci-e_scheme}, includes the following steps: 
\begin{enumerate}    
\item Source nodes perform feature extraction to obtain an embedding $\ev_i(t)$ of their view $\xv_i(t)$. Thus, the \emph{View Processing} module consists only of a single-view feature extraction network $\SingleViewBackbone$ and the intermediate representation $\zv_i(t)$ that each node sends to the central controller is a view embedding:
$\zv_i(t) {=} \ev_i(t)$; 
\item Each source node receives the current context $\cv(t)$ from the central controller, which consists of a view embedding  used to determine whether or not the current view should contribute to the classification task;
\item The \emph{Quality Estimation} module of each source node estimates how similar the embedding of its view is to the context by computing the cosine similarity between the two tensors: $Cosine(\cv(t), \ev_i(t)) {=} \frac{\cv(t){\cdot} \ev_i(t)}{\|\cv(t)\|\|\ev_i(t)\|}$. 
Then, the result is compared to a given  \emph{similarity threshold} $\gamma$ -- a configurable system parameter whose impact is analysed in Sec.~\ref{sub:threshold_eval}. Any view that is too similar to the context, i.e., exceeding the similarity threshold, will be discarded, while the others, for which $Cosine(\cv(t), \ev_i(t)) {<} \gamma$, will be forwarded to the central controller;

\item The central controller receives a collection of $V' {\leq} V$ view embeddings, which are aggregated by  the \emph{View Aggregator} module into a single embedding $\ev(t)$ using a view pooling function $\ViewPool$. Then, the controller completes the multi-view classification task by executing the classifier portion $\MultiViewHead$ of a multi-view CNN on $\ev(t)$, obtaining a prediction $\pred(t)$. The class prediction is  sent back to all source nodes;
\item As the information used as  current context is the pooled view embedding obtained at the previous time period  
(i.e., $ \cv(t) {=} \ev(t-1)$), 
 the \emph{Context Manager} module of the central controller will store the computed pooled embedding $\ev(t)$ and send it to each source node as context at the beginning of the next inference period.
\end{enumerate}

\begin{figure*}[t]
    \centering
    \begin{subfigure}[c]{0.48\textwidth}
        \centering
        \includegraphics[width=8.2cm]{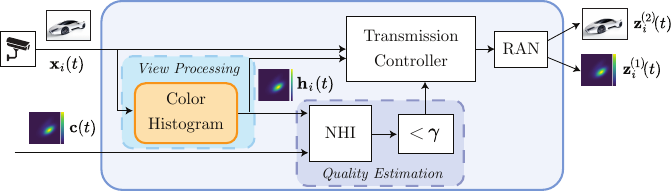}
        \caption{\label{fig:sci-ch_scheme_source_nodes}}
    \end{subfigure}
    \begin{subfigure}[c]{0.50\textwidth}
        \centering
        \includegraphics[width=8.4cm]{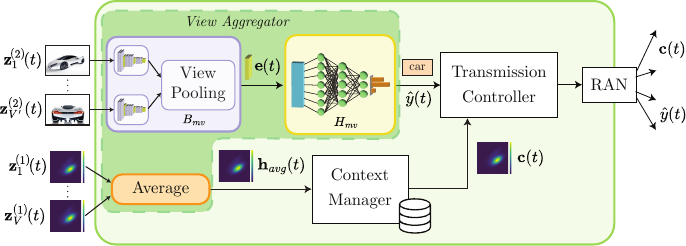}
        \caption{\label{fig:sci-ch_scheme_controller}}
    \end{subfigure}
    \caption{Schematic representation of the operations performed by a source node (a) and the central controller (b) in the selective centralized inference scheme based on color histograms (\SelectiveCentralizedInferenceColorHistTag{}).}
    \label{fig:sci-ch_scheme}
\end{figure*}

\paragraph{\textbf{\SelectiveCentralizedInferenceTag{} based on color histograms.}}
\label{par:sci-ch}
\begin{comment}
\paragraph{\textbf{Normalized Histogram Intersection.}}
Given two color histograms $\hv_1$ and $\hv_2$, along with $B$ their number of bins, we denote with $\text{NHI}$ the function responsible for computing the \emph{normalized histogram intersection} between $\hv_1$ and $\hv_2$:
\begin{equation*}
    NHI(\hv_1, \hv_2) = \frac{\sum_{j=1}^{B}\min(h_{1, j}, h_{2, j})}{\sum_{j=1}^{B}h_{2, j}}\,.
\end{equation*}
The result is a value in the range $[0,1]$ with $0$ meaning no intersection and $1$ meaning that $\hv_1$ is completely overlapped by $\hv_2$. 
\end{comment}

In this case, referred to as \SelectiveCentralizedInferenceColorHistTag, each source node decides whether or not to send its view to the central controller by comparing a light-weight descriptor of it, namely, a color histogram~\cite{manjunathColorTextureDescriptors2001}, to that of the other views captured within the {\em same time period}. 
%We use as light-weight image descriptors that source nodes can compute, share and use to compare their respective views. 
% REPETITION
% Specifically, each source node first computes the color histogram of its view, $hist(\xv_i(t))$, and sends it to the central controller. The latter averages the received histograms and sends the combined descriptor as context back  to source nodes. Each source node then decides whether or not to have its view contributing to the current inference task by calculating the similarity between the color histogram of its view and the average histogram received by the central controller, by comparing the result against a similarity threshold. We use the Normalized Histogram Intersection (NHI)~\cite{swainColorIndexing1991} as a light-weight measure to compute similarity between two color histograms. The NHI will return a value between 0 and 1, with the two values indicating null intersection and full overlap, respectively. 
The \SelectiveCentralizedInferenceColorHistTag{} scheme, depicted in Fig.~\ref{fig:sci-ch_scheme}, thus includes the following steps:
\begin{enumerate}
    \item Source nodes perform the color histogram computation with $B$ bins to obtain a light-weight descriptor $\hv_i(t)$ of their view $\xv_i(t)$. Therefore, the \emph{View Processing} module consists only of the color histogram computation function $hist$. 
    Each source node sends its color histogram $\hv_i(t)$ to the central controller as a first intermediate representation:
$
    \zv_i^{(1)}(t) {=} \hv_i(t)
    $;
    \item Each source node receives the current context $\cv(t)$ from the central controller. The context consists of an average color histogram $\hv_{avg}(t)$ that will be used to determine whether or not the current view should take part in the current classification task;
    \item The \emph{Quality Estimation} module of each source node estimates how similar the color histogram of its view is to the context. We use the Normalized Histogram Intersection (NHI)~\cite{swainColorIndexing1991} as a light-weight measure to compute similarity between two color histograms. The NHI will return a value between 0 and 1, with the two values indicating null intersection and full overlap, respectively. The result of NHI is then compared against a given similarity threshold $\gamma$ -- as for the previous case, the impact of the value of the similarity threshold will be determined as described in Sec.~\ref{sub:threshold_eval}. Only those views whose similarity to the context does not exceed $\gamma$, i.e., 
    $
    \textit{NHI}(\hv_i(t), \cv(t)) {<} \gamma
    $, will be forwarded (unprocessed) to the central controller:
    $
    \zv_i^{(2)}(t) {=} \xv_i(t)
    $;
    \item The \emph{View Aggregator} module of the central controller is responsible for two tasks. First, it aggregates the color histograms received from the source nodes into a single average descriptor $\hv_{avg}(t)$. Then, upon receiving a collection of $V' {\leq} V$ selected views, it executes a multi-view CNN model to obtain a class prediction $\pred(t)$. The class prediction is then sent back to all source nodes.
    \item The \emph{Context Manager} module sends the average color histogram $\hv_{avg}(t)$ to all source nodes as context:
    $
        \cv(t) {=} \hv_{avg}(t). 
    $
\end{enumerate}

\subsection{Ensemble inference schemes}
\label{sub:ensemble}

In ensemble inference schemes, each source node performs a local inference task based on its captured view. The class predictions resulting from these inferences are then aggregated by the central controller using a consensus protocol.

\begin{figure*}[t]
    \centering
    \begin{subfigure}[c]{0.51\textwidth}
        \centering
        \includegraphics[width=8.6cm]{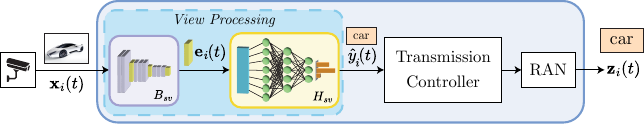}
        \caption{\label{fig:ei_scheme_source_nodes}}
    \end{subfigure}
    \begin{subfigure}[c]{0.47\textwidth}
        \centering
        \includegraphics[width=7.9cm]{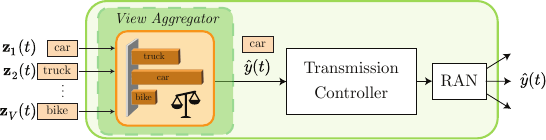}
        \caption{\label{fig:ei_scheme_controller}}
    \end{subfigure}
    \caption{Schematic representation of the operations performed by a source node (a) and the central controller (b) in the non-selective ensemble inference scheme (\EnsembleInferenceTag{}).}
    \label{fig:ei_scheme}
\end{figure*}

\subsubsection{Non-selective ensemble inference}
\label{sub:ei}

According to this scheme, named \EnsembleInferenceTag, single-view classification is performed locally by each source node while the central controller acts as a prediction aggregator. The \EnsembleInferenceTag{} scheme, depicted in Fig.~\ref{fig:ei_scheme}, consists of the following steps:
\begin{enumerate}
\item Source nodes perform local single-view classification on their input views $\xv_i(t)$; therefore the \emph{View Processing} module  implements a single-view CNN model to obtain a \emph{local prediction} $\pred_i(t)$. The local prediction is then forwarded to the central controller, i.e., 
$
\zv_i(t) {=} \pred_i(t)
$;
\item The central controller acts as a prediction aggregator, hence its \emph{View Aggregation} module implements a consensus protocol that selects as  final prediction the label appearing most frequently among the predictions made by source nodes. Thus, 
\[
\pred(t) = arg \max\limits_{y \in Y} \sum\limits_{i=1}^{} \mathcal{I}_y(\pred_i(t))
\]
where $\mathcal{I}_y$ for a given class label $y$ is the indicator function taking on 1 if $x {=} y$, and 0 otherwise. 
The final class prediction $\pred(t)$ is then sent back to the source nodes;
\item Being the scheme non-selective,  there is no sharing of contextual information between the central controller and source nodes. Consequently, it does require neither the \emph{Quality Estimation} at the source nodes nor the \emph{Context Manager} at the central controller.
\end{enumerate}

\subsubsection{Selective ensemble inference}
\label{sub:sei}
In \emph{selective ensemble inference schemes} (\SelectiveEnsembleInferenceTag) each source node decides whether or not to perform a local inference task and send its class prediction to the central controller, based on a measure of similarity between the current view and the context. 
%Only those views that yield sufficiently distinctive information with respect to the context will see their local prediction sent to the central controller to participate in the consensus protocol. 
Similarly to the selective centralized schemes described before, we investigate two variants of the \SelectiveEnsembleInferenceTag{} schemes: one based on \emph{view embeddings} and the other  on \emph{color histograms}.

\begin{figure*}[t]
    \centering
    \begin{subfigure}[c]{0.54\textwidth}
        \centering
        \includegraphics[width=9.1cm]{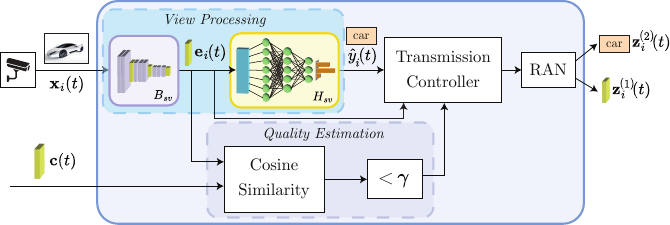}
        \caption{\label{fig:sei-e_scheme_source_nodes}}
    \end{subfigure}
    \begin{subfigure}[c]{0.44\textwidth}
        \centering
        \includegraphics[width=7.5cm]{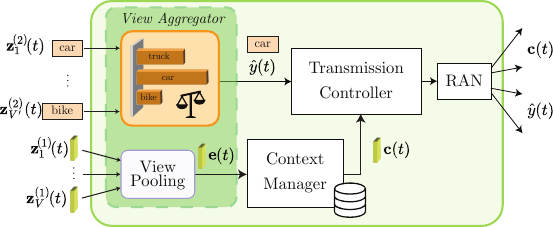}
        \caption{\label{fig:sei-e_scheme_controller}}
    \end{subfigure}
    \caption{Schematic representation of the operations performed by a source node (a) and the central controller (b) in the selective ensemble inference scheme based on embeddings (\SelectiveEnsembleInferenceEmbeddingTag{}).}
    \label{fig:sei-e_scheme}
\end{figure*}

\paragraph{\textbf{\SelectiveEnsembleInferenceTag{} based on embeddings.}}
\label{par:sei-e}

In this case, referred to as \mbox{\SelectiveEnsembleInferenceEmbeddingTag}, %each source node decides whether or not to perform its local inference by comparing the embedding of the current view to a representation of the features collectively captured by the network at the {\em previous time period}. 
we use the pooled view embedding computed by the network at the previous time period as contextual information to discriminate views. Such a context is computed by the central controller and disseminated to source nodes in the same way described for the \SelectiveCentralizedInferenceEmbeddingTag{} scheme. Differently from the corresponding centralized case, the context here is used to decide whether or not each source node should perform its local inference.

%Before performing its local inference, each source node estimates how different its locally-captured view is from the context by computing the cosine similarity between the view embedding and the context. Views that fail to convey sufficiently distinctive information with respect to the context (as determined by the similarity threshold) will be discarded. If a view is processed, then the resulting class prediction will be sent to the central controller (along with the corresponding embedding for context aggregation) to participate in the consensus protocol.
Specifically, the \SelectiveEnsembleInferenceEmbeddingTag{} scheme, depicted in Fig.~\ref{fig:sei-e_scheme}, consists of the following  steps:
\begin{enumerate}
\item Source nodes perform feature extraction to obtain an embedding $\ev_i(t)$ of their view $\xv_i(t)$. The \emph{View Processing} module consists of a single-view CNN model in which the feature extraction part $\SingleViewBackbone$ is always executed whereas the classification part $\SingleViewHead$ is executed only conditionally. Each source node sends its embedding $\ev_i(t)$ to the central controller as a first intermediate representation:
$
\zv_i^{(1)}(t) {=} \ev_i(t)$.
Upon receiving the current context and if its local view passes the quality estimation check, the source node  completes the local inference task by executing $\SingleViewHead$ to produce a local class prediction $\pred_i(t)$. The local class prediction is then sent to the central controller to participate in the consensus protocol:
$
\zv_i^{(2)}(t) {=} \pred_i(t)$;
\item Each source node receives the current context $\cv(t)$ from the central controller. The context consists of a view embedding that is used to determine whether or not the current view should take part in the classification task;

\item The \emph{Quality Estimation} module of each source node estimates how similar the embedding of its view is to the context by computing the cosine similarity between the two tensors.  The result is then compared to a given similarity threshold $\gamma$. Only those embeddings whose similarity to the context does not exceed $\gamma$, i.e., $Cosine(\cv(t), \ev_i(t)) {<} \gamma$, will be used to derive a local prediction;

\item The \emph{View Aggregator} module of the central controller is responsible for two tasks. First, it aggregates the view embeddings received from the source nodes into a pooled embedding $\ev(t)$ using a view pooling function $\ViewPool$. Then, upon receiving a collection of $V' {\leq} V$ local predictions, it executes a consensus protocol selecting as a final prediction the most frequent label among them. The final class prediction $\pred(t)$ will then be sent back to source nodes;

\item The information used as the current context $\cv(t)$ is the pooled view embedding obtained at the previous time period; i.e., 
$\cv(t) {=} \ev(t-1)$.
Therefore, the \emph{Context Manager} module of the central controller will store the computed pooled embedding $\ev(t)$ and send it to each source node as context at the beginning of the next time period.
\end{enumerate}

\begin{figure*}[t]
    \centering
    \begin{subfigure}[c]{0.52\textwidth}
        \centering
        \includegraphics[width=8.8cm]{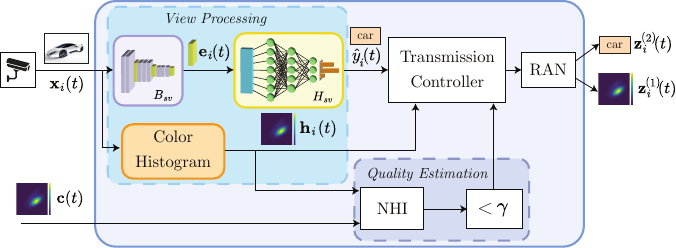}
        \caption{\label{fig:sei-ch_scheme_source_nodes}}
    \end{subfigure}
    \begin{subfigure}[c]{0.46\textwidth}
        \centering
        \includegraphics[width=7.8cm]{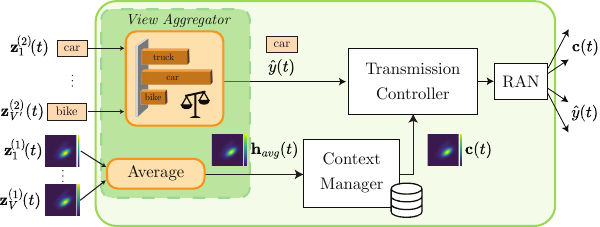}
        \caption{\label{fig:sei-ch_scheme_controller}}
    \end{subfigure}
    \caption{Schematic representation of the operations performed by a source node (a) and the central controller (b) in the selective ensemble inference scheme based on color histograms (\SelectiveEnsembleInferenceColorHistTag{}).}
    \label{fig:sei-ch_scheme}
\end{figure*}

\paragraph{\textbf{\SelectiveEnsembleInferenceTag{} based on color histograms.}}
\label{par:sei-ch}

In this case, referred to as \SelectiveEnsembleInferenceColorHistTag, we use color histograms as contextual information representing the views collectively captured at the {\em current time period}. Color histograms are computed by source nodes and aggregated by the central controller in the same way as described for the \SelectiveCentralizedInferenceColorHistTag{} scheme. Unlike in the corresponding centralized case, the context is used to decide whether or not each source node should perform its local inference.
%Before performing its local inference, each source node estimates the similarity of its view to the others by computing the NHI between the color histogram of the local view and the context. Views that fail to convey sufficiently distinctive information with respect to the context (as determined by the similarity threshold) will be discarded. 
The \SelectiveEnsembleInferenceColorHistTag{} scheme, depicted in Fig.~\ref{fig:sei-ch_scheme},  thus operates as follows: 
\begin{enumerate}
    \item The \emph{View Processing} module of source nodes consists of the color histogram computation function $hist$ and a single-view CNN model. First, each source node computes the color histogram (with $B$ bins) of its view $\xv_i(t)$ to obtain a lightweight descriptor $\hv_i(t)$. The color histogram is sent to the central controller as a first intermediate representation: $ \zv_i^{(1)}(t) {=} \hv_i(t)$.
  Upon receiving the context and if its local view passes the quality estimation check, the source node  executes the single-view CNN to produce the local prediction $\pred_i(t)$, which is then sent to the central controller to participate in the consensus protocol:
 $ \zv_i^{(2)}(t) {=} \pred_i(t)$;
 
    \item Each source node receives the current context $\cv(t)$ from the central controller. The context consists of an average color histogram $\hv_{avg}(t)$, that is used to determine whether or not to perform the local inference task;
    
    \item The \emph{Quality Estimation} module of each source node estimates how similar the color histogram of its view is to the context by computing the NHI between the two histograms. The result is compared to a given similarity threshold $\gamma$: only those views whose similarity to the context does not exceed $\gamma$ (i.e., s.t.\ $\text{NHI}(\hv_i(t), \cv(t)) {<} \gamma$) are subject to local inference and participate in the consensus protocol;
    \item The \emph{View Aggregator} module of the central controller is responsible for two tasks. First, it aggregates the color histograms received from the source nodes into a single average descriptor $\hv_{avg}(t)$. Upon receiving a collection of $V' {\leq} V$ local predictions, it executes the consensus protocol selecting as a final prediction the most frequent label among them. The final class prediction $\pred(t)$ will then be sent back to source nodes; 
    \item The \emph{Context Manager} module sends the average color histogram $\hv_{avg}(t)$ to all source nodes as context:
    $\cv(t) {=} \hv_{avg}(t)$.
\end{enumerate} 

\section{Experimental Evaluation}
\label{sec:exp}
This section presents our experimental setup (Sec.~\ref{sub:setup}) and the obtained results. In particular, we first analyze the impact of different values of the similarity threshold used  in the selective schemes (Sec.~\ref{sub:threshold_eval}). 
Then %{\color{red} \sout{we evaluate how the accuracy and communication overhead of the inference schemes we consider is affected by different networking and data-related conditions. (Sec.~\ref{sub:comparison})}}  
we compare the different schemes in terms of accuracy, communication overhead, and inference latency, under different operational  conditions (Sec.~\ref{sub:comparison}).

\subsection{Experimental setup}
\label{sub:setup} 
Experiments are performed by running pre-trained single- and multi-view models under the experimental conditions detailed below.

{\bf Classification dataset.} 
We use the ModelNet40~\cite{zhirong3DShapeNets2015} dataset to train the underlying CNN models and  test the different collaborative inference schemes. The dataset consists of a collection of multi-view 3D shape recognition instances with 12 views each and labels spanning across 40 classes. Instances are split into 9,483 training instances and 2,468 test instances. Consistently with previous works~\cite{suMultiviewConvolutionalNeural2015, suDeeperLookAt3DShape2018}, we use the standard train-test split provided by the dataset in our experiments. At a given time period $t$, the selective inference schemes based on view embeddings (\SelectiveCentralizedInferenceEmbeddingTag{} and \SelectiveEnsembleInferenceEmbeddingTag{}) require the embeddings of the views captured at $t-1$ to generate the contextual information used to discriminate views. 
As the instances in our dataset are not temporally correlated with each other, for each instance we randomly select 6 out of the 12 available views and use them as the set of views captured by the network at the previous time period to derive the context. The remaining 6 views are used as the multi-view collection captured at the current time period. For a fair comparison, we limit the maximum number of views used for each instance to 6 by randomly sampling each multi-view collection in the test set. 

{\bf Models implementation and training.} 
We implemented each scheme as a standalone deep learning model using PyTorch. Centralized inference schemes were implemented by instrumenting a pre-trained multi-view CNN ({\emph{base multi-view model}), whereas ensemble inference schemes were implemented by extending a pre-trained single-view CNN model (\emph{base single-view model}). The same base model was used to implement all schemes belonging to the same category. First, we trained the base single-view model $\SingleViewCNN$, starting from a general-purpose VGG-16~\cite{simonyanVeryDeepConvolutional2015} image classification network, pre-trained on ImageNet~\cite{dengImagenet2009}. The VGG-16 model consists of five convolutional blocks for feature extraction, followed by three fully-connected layers and a softmax layer for classification. The default input size for the model is 224$\times$224 pixels. This general-purpose network was then fine-tuned on the single-view images of the ModelNet40 dataset. Training was performed for 30 epochs, using the Adam optimizer with a learning rate of $5 {\times} 10^{-5}$ and $64$ instances per batch over the entire training set. The base multi-view model $\MultiViewCNN$ was then built using the base single-view network $\SingleViewCNN$ as a backbone, as described in Sec.~\ref{sub:problem}. The multi-view model was trained to jointly classify all the views of the multi-view collections in the training set of ModelNet40. Training was performed for 30 epochs, using the Adam optimizer with a learning rate of $1 \times 10^{-5}$, and $8$ multi-view instances per batch over the entire training set.

{\bf Hardware description.}
Experiments were run on Google Colab's virtual machines using a Pro+ subscription. The hardware specifications for the used virtual machines consists of 8 vCPUs Intel(R) Xeon(R) @ 2.00\,GHz, 52\,GB of RAM, an NVIDIA Tesla V100 GPU with 16\,GB of GPU memory, 170\,GB of persistent disk storage and a 1\, Gbps-network connection.

{\bf Evaluation metrics.}
Through our experiments, we evaluate prediction accuracy and communication overhead for the different collaborative schemes under various network and data-related settings.
Accuracy is obtained directly by evaluating the models over our test set. As for the communication overhead,  
%{\color{red} \sout{we compute it by considering the number and size of the messages exchanged between the system nodes according to the considered inference task.}}
we consider a reference scenario in which  nodes communicate through a TCP connection over a 5G link. Table~\ref{tab:network_scenario} presents the 5G parameter settings used. We then compute the communication  overhead for each scheme as the total number  of  transmitted bytes per inference, taking into account: the number of messages exchanged between the nodes, their respective size at the application layer (presented in Table~\ref{tab:packet_size}), all protocol headers and the size of control packets.
For each scheme, we also report the {\em transmission gain}, indicating the decrease in the percentage of  transmitted views under each scheme with respect to a given baseline (that will be defined for each experiment).
Inference latency, accounting for both data transfer and data processing, is also computed considering the aforementioned reference scenario.
%{\color{red} \sout{We report the communication overhead both in terms of {\em transmission gain}, indicating the decrease in the percentage of transmitted messages under each scheme against a baseline (that will be defined for each experiment) and in terms of average number of {\em transmitted bytes}  per instance. To compute the latter, we consider a standard uncompressed binary encoding of the tensors shared between nodes.}}

\begin{table}
\centering
\caption{Reference scenario: radio channel parameters.}
\label{tab:network_scenario}

\begin{tblr}{
  cell{1}{1} = {c=2}{},
  cell{2}{1} = {c=2}{},
  cell{3}{1} = {c=2}{},
  cell{4}{1} = {c=2}{},
  cell{5}{1} = {c=2}{},
  cell{6}{1} = {c=2}{},
}
\textbf{Parameter}                 &           & \textbf{Value}  \\
\hline
Carrier Frequency            &           & 3.5 GHz (FR1)    \\
Channel Bandwidth         &           & 50 MHz \\
Subcarrier Spacing (SCS) &           & 15 kHz \\
No. of MIMO Layers               &           & 2      \\
Resource Blocks allocated to 5G slice              &           & 50 
\end{tblr}
\end{table}

\begin{table}
\centering
\caption{Size of  exchanged application-layer packets for each scheme}
\label{tab:packet_size}

\begin{tblr}{
  cells = {c},
  cell{2}{1} = {r=2}{},
  cell{4}{1} = {r=3}{},
  cell{7}{1} = {r=4}{},
  cell{11}{1} = {r=2}{},
  cell{13}{1} = {r=4}{},
  cell{17}{1} = {r=4}{},
  hline{2,4,7,11,13,17} = {-}{},
}
\textbf{Scheme} & \textbf{Message description}      & \textbf{Size [bytes]} \\ 
CI     & View message $\zv_i(t)$             & 602,112 \\
       & Prediction message $\pred(t)$       & 1     \\
SCI-E  & Context message $\cv(t)$          & 100,352 \\
       & Embedding message $\zv_i(t)$        & 100,352 \\
       & Prediction message $\pred(t)$       & 1     \\
SCI-CH & Color histogram message $\zv_i^{(1)}(t)$  & 4,096   \\
       & Context message $\cv(t)$          & 4,096   \\
       & View message $\zv_i^{(2)}(t)$             & 602,112 \\
       & Prediction message $\pred(t)$       & 1     \\
EI     & Prediction message $\zv_i(t)$       & 1     \\
       & Final prediction message $\pred(t)$ & 1     \\
SEI-E  & Context message $\cv(t)$          & 100,352 \\
       & Embedding message $\zv_i^{(2)}(t)$        & 100,352 \\
       & Prediction message $\zv_i^{(1)}(t)$       & 1     \\
       & Final prediction message $\pred(t)$ & 1     \\
SEI-CH & Color histogram message $\zv_i^{(1)}(t)$  & 4,096   \\
       & Context message $\cv(t)$          & 4,096   \\
       & Prediction message $\zv_i^{(2)}(t)$       & 1     \\
       & Final prediction message $\pred(t)$ & 1     
\end{tblr}
\end{table}

% \begin{table}
% \centering
% \caption{{\color{blue}Size of  application-layer packets exchanged between nodes}}
% {\color{blue}
% \begin{tabular}{ccc}
% \textbf{Message Content}  & \textbf{Tensor Size}   & \textbf{Packet Size [bytes]}  \\
% \hline
% View             & 224 $\times$ 224 $\times$ 3 & 602,112                            \\
% View Embedding   & 7 $\times$ 7 $\times$ 512   & 100,352                            \\
% Color Histogram  & 32 $\times$ 32                               & 4,096                              \\
% Inference Result & 1                                             & 1                               
% \end{tabular}
% }
% \end{table}

\subsection{Similarity thresholds evaluation}
\label{sub:threshold_eval}

%Selective inference schemes rely on a pre-determined similarity threshold parameter $\gamma$ to discriminate whether a view should participate in the current inference or not. 
Here, we provide an experimental evaluation assessing the impact on accuracy and communication overhead of different values of similarity threshold $\gamma$ for different variants of the selective inference schemes, i.e., embedding-based and color histogram-based. For brevity, we only show the results obtained for  selective centralized schemes, as we observed a very similar behavior in the case of selective ensemble approaches.  

{\bf Threshold analysis for embedding-based schemes.} 
We evaluate the \SelectiveCentralizedInferenceEmbeddingTag{} scheme for different threshold values, namely, in the range $[0.1,0.9]$ with 0.1 increments and a growing number of source nodes from $1$ to $6$. As mentioned above, the remaining $6$ views are used to represent the context. For each pair of  $(\gamma, N)$ values, we performed 12  evaluations of the \SelectiveCentralizedInferenceEmbeddingTag{} scheme over the entire test set, each time selecting different random subsets of views.  Fig.~\ref{fig:emb_threshold} shows the obtained results, averaged over the 12 runs, in terms of accuracy (a),  transmission gain (due to discarded views) (b), and 
%{\color{red}\sout{amount of transmitted data}} 
communication overhead per inference (c). The \SelectiveCentralizedInferenceEmbeddingTag{} scheme with $\gamma {=} 1$, in which no view gets discarded, is used as a baseline. We only report the results obtained for $\gamma {\in} \{0.3, 0.4, 0.5, 0.6\}$ to reduce visual clutter in the plots. Any $\gamma {>} 0.6$ would further degrade accuracy, and any $\gamma {<} 0.3$ would be virtually indistinguishable from the baseline.

\begin{figure*}[!htp]
\centering
\includegraphics[width=\textwidth]{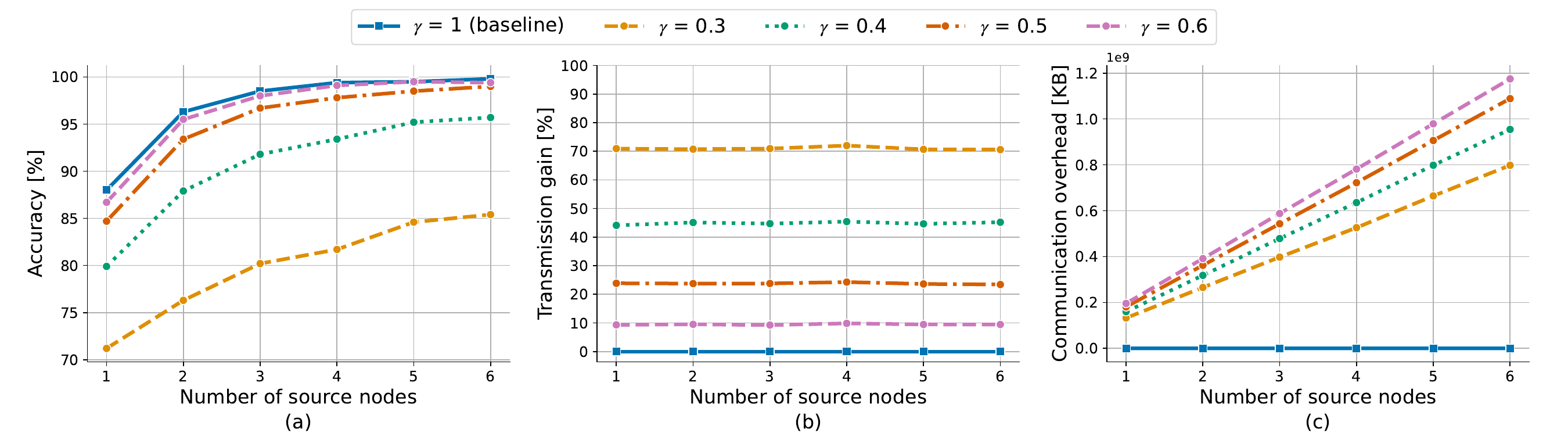}
\caption{Impact of different similarity threshold values for the \SelectiveCentralizedInferenceEmbeddingTag{} scheme with increasing number of source nodes: accuracy (a), transmission gain (b), and the communication overhead  per inference  (c).}
\label{fig:emb_threshold}
\end{figure*}

\begin{figure*}[!htp]
\centering
\includegraphics[width=\textwidth]{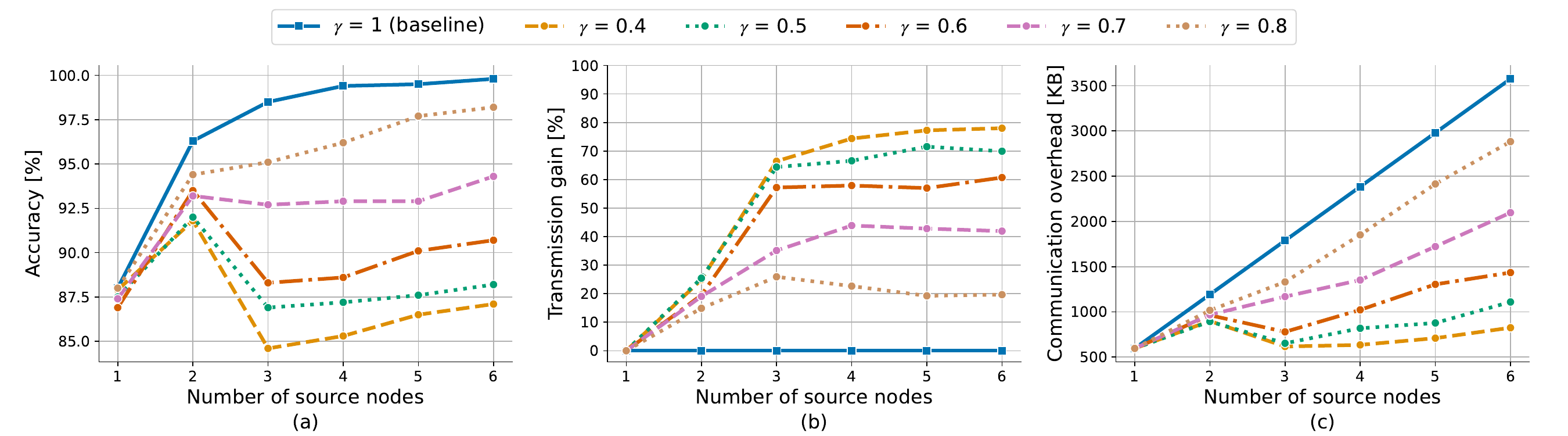}
\caption{Impact of different similarity threshold values for the \SelectiveCentralizedInferenceColorHistTag{} scheme with increasing number of source nodes: accuracy (a), transmission gain (b), and the communication overhead per inference  (c).}
\label{fig:hist_threshold}
\end{figure*}

The plots show that the transmission gain, and thus the ratio of discarded views, is independent of the number of source nodes. When only one source node is available, discarding a view corresponds to dropping the current inference task.
Even if the transmission gains vary widely for the different values of $\gamma$ (ranging from 10\% to 70\%), the difference in  
%{\color{red}\sout{the amount of data transmitted}} 
communication overhead per inference is more limited (ranging from 4\% to 35\% concerning the baseline).  
As expected, the accuracy decreases as the number of discarded views increases. Overall, however, the system is still able to reach high accuracy (above 90\%) while leveraging the benefit of sizable transmission gains (between 24\% and 44\%) for values of $\gamma$ between 0.4 and 0.5. Importantly, the plots highlight that a similarity threshold of 0.4 achieves the best trade-off in terms of accuracy and transmission efficiency, with a 44\% reduction in the number of transmitted views while degrading accuracy by just 4.4\% on average, with respect to the baseline. 

{\bf Threshold analysis for color histogram-based schemes.} We perform the same threshold analysis on the \SelectiveCentralizedInferenceColorHistTag{} scheme.
Fig.~\ref{fig:hist_threshold} presents the results, again, in terms of accuracy (a),  transmission gain (b), and  
%{\color{red}\sout{amount of transmitted data}} 
communication overhead per inference (c). 
The \SelectiveCentralizedInferenceColorHistTag{} scheme with $\gamma {=} 1$, in which no view gets discarded, is used as a baseline, and, for brevity, we only report the results obtained for $\gamma {\in} \{0.4, 0.5, 0.6, 0.7, 0.8\}$ as the other values do not provide additional insights.

Consistently with the intuition, accuracy decreases as the value of $\gamma$ gets smaller, and thus more views are discarded. Interestingly, for the strategies that most aggressively discard views ($\gamma{<}0.7$), the accuracy follows a non-monotone trend. It spikes when moving from a single source node to two, capitalizing on the benefits of multi-view classification as the ratio of discarded views is still reasonably low (below 26\%). With three source nodes, the ratio of discarded views has its steepest ascent, resulting in a sudden drop in accuracy. This is due to the average color histogram, used by \SelectiveCentralizedInferenceColorHistTag{} as a context, which is very sensitive to the variability in the color distribution of the individual views when computed on a small number of sources. This fact, paired with an overly selective threshold, may lead the scheme to discard some informative views whose contribution is crucial to achieving better accuracy when the total number of views is scarce. As more views become available, the average color histogram becomes an adequate descriptor of the collection of views captured by the network, and the downside of discarding some of the informative views becomes less severe.

Differently from the embedding-based case, the ratio of discarded views in the histogram case grows with the number of available nodes. When only one source node is available, no view gets discarded and each inference task is reduced to single-view classification. Note that no inference task is dropped in the \SelectiveCentralizedInferenceColorHistTag{} scheme. As the number of available views increases, so does the transmission gain, growing rapidly at first until reaching a plateau at around 5 views. 
A similarity threshold of 0.7 seems to achieve the best trade-off between accuracy and transmission gain, attaining at least 92.5\% accuracy while discarding between 34\% and 52\% of the views.

{\bf Lessons learnt on the impact of similarity thresholds. }
Selecting a similarity threshold for each of the proposed selective inference schemes is crucial in determining the trade-off between accuracy and communication overhead the network achieves.
The value of $\gamma$ can be dynamically tuned to adapt to different network loads, as discarding views more aggressively allows the system to decrease bandwidth utilization at the cost of a loss in accuracy, with embedding-based schemes potentially experiencing a significantly greater degradation than color histogram-based schemes.

\subsection{Comparative analysis}
\label{sub:comparison}

In this section, we provide a 
quantitative comparison of the schemes in terms of prediction accuracy, communication overhead,  and inference latency. For each scheme, we perform multiple experiments considering a number $N$ of source nodes ranging from $1$ to $6$. In each experiment, the  specified number of views are randomly selected among the $12$ available. We repeat each experiment over $12$ runs and  average the results. For the selective schemes, we use the value of the similarity threshold $\gamma$ that, based on our experiments, yields the best trade-off between accuracy and transmission gain, i.e., $\gamma {=} 0.4$ for embedding-based schemes and $\gamma {=} 0.7$ for color histogram-based schemes. Results are reported in Table~\ref{tab:exp} as well as in Fig.~\ref{fig:robustness}, in terms of accuracy (a), transmission gain (b), communication overhead (c), and inference latency (d). The \CentralizedInferenceTag{} scheme is used as a baseline. 
Then, we investigate how the accuracy of the schemes is affected by different networking and data-related conditions, such as partial availability of the source nodes and noise in the input data. 
Finally, Table~\ref{table:properties} presents a qualitative comparison of the proposed schemes.

\paragraph{\textbf{Accuracy and communication  overhead trade-off.}}
\label{subsub:tradeoff}
%We start by varying the number of source nodes $N$ and evaluate each scheme over multiple test sets, each comprising a restricted number of views to represent different numbers of source nodes.
%For each $N{\in}\{1, {\lots}, 6\}$, we performed $12$ distinct runs, every time selecting a different random subset of views. The results are then averaged over the $12$ runs. 

Comparing the accuracy of the different schemes, we observe that the ensemble schemes, enabling local inference at the source nodes,  provide good performance (namely, between 71.92\% and 83.75\% accuracy in average, see Table~\ref{tab:exp}), while reducing the 
%{\color{red}\sout{amount of transmitted data}} 
communication overhead by one or more orders of magnitude.    
Conversely, the \CentralizedInferenceTag{} scheme exhibits slightly higher accuracy but leads to the largest bandwidth consumption.  
%The rest of the schemes achieve different trade-offs between accuracy and communication overhead. 
Furthermore, the \SelectiveCentralizedInferenceEmbeddingTag{} and \SelectiveCentralizedInferenceColorHistTag{} schemes attain comparable accuracy levels and transmission gains when at least $3$ source nodes are available, with the former scheme slightly outperforming the latter.  
Overall, the results highlight the benefits of selective schemes, which still provide considerably accurate predictions (above 92\% accuracy) while reducing the 
%{\color{red}\sout{amount of transmitted data}} 
communication overhead from 18\% to 74\% (on average) with respect to the non-selective case, thanks to their ability to discard redundant views. 

\begin{table*}
\caption{Results of the comparison among inference schemes as the number of source nodes varies.}
\label{tab:exp}
\resizebox{\linewidth}{!}{
\hfill%
\begin{subtable}[t]{0.56\textwidth}
    \begin{tabular}{c | c c c c c c}
        \multicolumn{7}{c}{Accuracy [\%]} \\
        \cline{1-7}
        N & CI & EI & SCI-E & SCI-CH & SEI-E & SEI-CH \\
        \hline
       1 & 88.00 & 76.80 & 79.90 & 87.40 & 67.30 & 76.20 \\
       2 & 96.30 & 78.00 & 87.90 & 93.20 & 67.90 & 78.50 \\
       3 & 98.50 & 83.80 & 91.80 & 92.70 & 71.50 & 79.60  \\
       4 & 99.40 & 87.20 & 93.40 & 92.90 & 73.20 & 79.70  \\
       5 & 99.50 & 87.70 & 95.20 & 92.90 & 75.30 & 80.70  \\
       6 & 99.80 & 89.00 & 95.74 & 94.30 & 76.30 & 82.70  \\
       \hline
       Mean & 96.92 & 83.75 & 90.66 & 92.23 & 71.92 & 79.57 \\
       Std Dev & 4.55 & 5.22 & 5.98 & 2.44 & 3.74 & 2.17 \\
    \end{tabular}
\end{subtable}
\hfill%
\begin{subtable}[t]{0.56\textwidth}
    \begin{tabular}{c | c c c c c c }
        \multicolumn{7}{c}{Transmission Gain [\%]} \\
        \cline{1-7}
        N & CI & EI & SCI-E & SCI-CH & SEI-E & SEI-CH \\
        \hline
       1 & 0.00 & 0.00 & 44.12 & 0.00 & 53.99 & 0.00    \\
       2 & 0.00 & 0.00 & 45.10 & 19.04 & 55.91 & 19.45 \\
       3 & 0.00 & 0.00 & 44.70 & 35.12 & 54.84 & 36.49  \\
       4 & 0.00 & 0.00 & 45.42 & 43.88 & 55.06 & 43.36  \\
       5 & 0.00 & 0.00 & 44.63 & 42.80 & 54.52 & 43.16  \\
       6 & 0.00 & 0.00 & 45.20 & 41.93 & 54.40 & 40.64  \\
       \hline
       Mean    & 0.00 & 0.00 & 44.86 & 30.46 & 54.79 & 30.52  \\
       Std Dev & 0.00 & 0.00 & 0.47 & 17.57 & 0.66 & 17.42   \\
    \end{tabular}
\end{subtable}
\hfill%
}
\\[2ex]
\resizebox{\linewidth}{!}{
\hfill%
\begin{subtable}[t]{0.62\textwidth}
    \begin{tabular}{c | c c c c c c }
        \multicolumn{7}{c}{Communication Overhead [KB]} \\
        \cline{1-7}
        N & CI & EI & SCI-E & SCI-CH & SEI-E & SEI-CH \\
        \hline
        1       & 623.58  & 0.17 & 207.95 & 632.09  & 208.03 & 8.69   \\
        2       & 1247.16 & 0.35 & 353.84 & 1026.77 & 342.37 & 17.31  \\
        3       & 1870.74 & 0.52 & 507.97 & 1239.36 & 485.33 & 25.88  \\
        4       & 2494.32 & 0.70 & 662.14 & 1433.82 & 629.43 & 34.46  \\
        5       & 3117.90 & 0.87 & 823.22 & 1825.98 & 779.07 & 43.07  \\
        6       & 3741.47 & 1.04 & 979.50 & 2223.69 & 927.96 & 51.71  \\
        \hline
        Mean    & 2182.53 & 0.61 & 589.10 & 1396.95 & 562.03 & 30.19  \\
        Std Dev & 1166.61 & 0.33 & 289.75 & 568.47  & 270.18 & 16.09  
    \end{tabular}
\end{subtable}
% \begin{subtable}[t]{0.62\textwidth}
%     \begin{tabular}{c | c c c c c c }
%         \multicolumn{7}{c}{Communication Overhead [KB]} \\
%         \cline{1-7}
%         N & CI & EI & SCI-E & SCI-CH & SEI-E & SEI-CH \\
%         \hline 
%        1 & 588.001 & 0.002 & 152.762 & 596.001 & 143.090 & 8.002 \\
%        2 & 1176.002 & 0.004 & 303.611 & 968.127 & 282.428 & 16.004 \\
%        3 & 1764.003 & 0.006 & 456.572 & 1168.557 & 426.762 & 24.005 \\
%        4 & 2352.004 & 0.008 & 605.966 & 1351.899 & 568.158 & 32.006 \\
%        5 & 2940.005 & 0.010 & 761.338 & 1721.655 & 712.837 & 40.008 \\
%        6 & 3528.016 & 0.012 & 910.238 & 2096.645 & 856.119 & 48.009 \\
%        \hline
%        Mean & 2058.003 & 0.007 & 531.748 & 1317.147 & 498.232 & 28.006 \\
%        Std Dev & 1100.049 & 0.004 & 283.832 & 535.985 & 267.145 & 14.969 \\
%     \end{tabular}
% \end{subtable}
\hfill%
\begin{subtable}[t]{0.62\textwidth}
    \begin{tabular}{c | c c c c c c }
        \multicolumn{7}{c}{Inference Latency [ms]} \\
        \cline{1-7}
        N & CI & EI & SCI-E & SCI-CH & SEI-E & SEI-CH \\
        \hline
        1       & 69.15  & 20.46 & 40.49  & 73.46  & 43.19  & 24.77  \\
        2       & 140.37 & 20.48 & 64.29  & 145.59 & 66.95  & 25.71  \\
        3       & 216.36 & 20.51 & 89.68  & 222.55 & 92.30  & 26.70  \\
        4       & 281.86 & 20.52 & 111.57 & 288.89 & 114.15 & 27.56  \\
        5       & 332.99 & 20.54 & 128.66 & 340.68 & 131.21 & 28.23  \\
        6       & 415.36 & 20.56 & 156.18 & 424.11 & 158.69 & 29.31  \\
        \hline
        Mean    & 242.68 & 20.51 & 98.48  & 249.21 & 101.08 & 27.05  \\
        Std Dev & 127.11 & 0.04  & 42.48  & 128.74 & 42.40  & 1.67   
    \end{tabular}
\end{subtable}
\hfill%
}
\end{table*}

\begin{figure*}
\centering
\includegraphics[width=\textwidth]{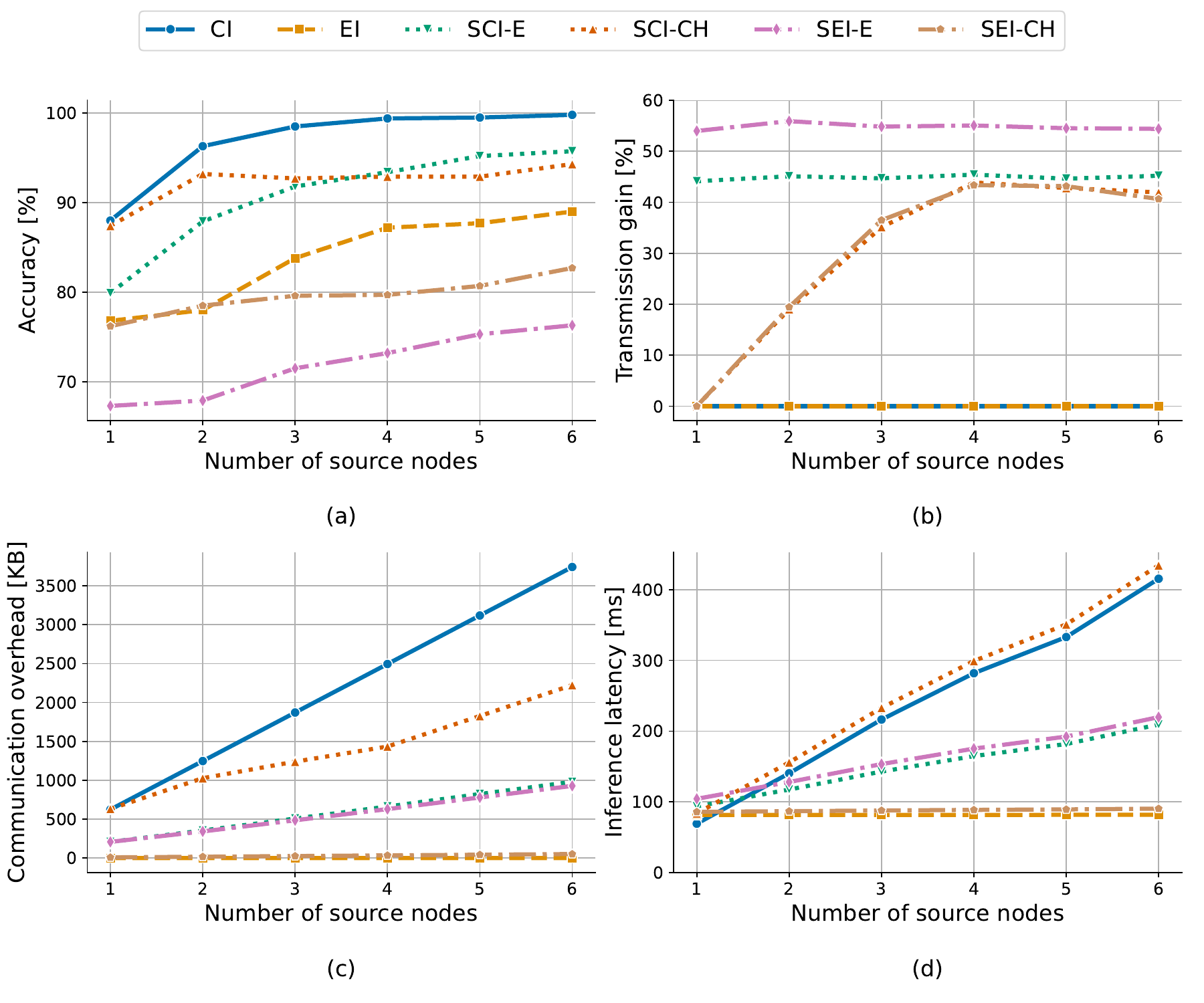}
\caption{Comparison in terms of accuracy (a), percentage of discarded views (b), communication overhead per inference (c), and inference latency (d) among the different schemes for increasing the number of source nodes. }\label{fig:robustness}
\end{figure*}
 
When considering bandwidth utilization per inference, the centralized selective scheme based on embeddings emerges as a clear winner with respect to the one based on color histograms, given the difference in the size of the intermediate representations exchanged between the source nodes and the controller in the two cases. However,  embedding-based selective schemes require, in general, source nodes to sustain a higher computational burden and, thus, are only feasible in scenarios in which devices are equipped with sufficient computing capability.  Finally, considering the ensemble approach, selective schemes always perform worse than non-selective ones, both in terms of accuracy and communication overhead.

{\bf Lessons learnt on accuracy-communication overhead trade-off.}
Ensemble schemes lead to a limited decrease in accuracy relative to centralized schemes while greatly reducing the communication overhead, making them more suitable in network-constrained scenarios. On the contrary, schemes in which nodes exchange raw images would, in general, entail a higher communication overhead. The communication burden decreases as the representation of the exchanged views becomes more refined (and lower-dimensional). Different trade-offs can be explored by applying (lossy) feature compression techniques before view transmission. Overall, selective schemes effectively reduce the required communication overhead with respect to their non-selective counterparts, sacrificing little accuracy. Furthermore, they can be tuned to achieve different trade-offs between accuracy and communication overhead by varying the similarity threshold parameter $\gamma$.  
%As a result, these schemes are particularly suited for scenarios in which the availability of network resources changes dynamically, decreasing $\gamma$, and thus reducing the number of transmitted views when the network is congested, or increasing it, and thus sending more views when enough bandwidth is available.  

Regarding accuracy, we take as base case a large multi-view CNN executed on a single machine. The \CentralizedInferenceTag{} scheme, in which all available views are fed unprocessed to a single large MVCNN running at the edge, is indeed the implementation achieving the highest accuracy. The accuracy degrades as the model gets split, and possibly compressed.  Comparing  \SelectiveEnsembleInferenceEmbeddingTag{} and \SelectiveEnsembleInferenceColorHistTag{}, the former yields 6\% decrease in accuracy, but 14\% increase in transmission gain.  For the centralized schemes,   \SelectiveCentralizedInferenceEmbeddingTag{} provides the best trade-off between accuracy and communication overhead, while  \SelectiveCentralizedInferenceColorHistTag{} is more versatile, requiring far lower computational resources at source nodes and reaching comparable levels of accuracy at the cost of a higher bandwidth utilization.
%This is due to the fact that the more the computation  load is moved to the source nodes,  the higher the level of refinement (embeddings or predictions instead of raw images) at which data are aggregated and then used to get the inference outcome. In addition, non-selective schemes, in which the system uses all data at its disposal for each inference, are  more accurate than the selective ones.  
However, our results show that, whenever a limited accuracy degradation is allowed, ensemble schemes can provide an excellent trade-off between accuracy and communication overhead, and selective schemes are very effective in further reducing the latter. 
%It follows that, if the application scenario requires very high level of accuracy, the \CentralizedInferenceTag{} scheme or a selective centralized scheme with a mildly aggressive threshold will be more suited. 

\paragraph{\textbf{Inference latency.}}
\label{par:latency}

To estimate the inference latency of the different schemes, we consider the previously mentioned reference scenario with 6 source nodes,  randomly located  within the coverage of a 5G base station (gNB), resulting in  levels of signal-to-noise ratio (SNR) between 0 and 20\,dB.  %Table~\ref{tab:channel_quality} describes the signal quality parameters assumed for each UE and the corresponding Modulation Coding Scheme. 
We recall that we consider a slice of $50$ Resource Blocks (RBs) allocated for the service. In the following experiments, the available RBs are evenly assigned to the source nodes. 

For each scheme, we estimate inference latency by considering two main components: the transmission and the processing latency. %Transmission delays are estimated by taking into account the size of each packet to be sent (as computed for the communication overhead) and the throughput of the channel. 
Notably, the processing  latency is  computed by profiling the execution time of the processing modules foreseen by  each scheme and using a virtual machine equipped with a single NVIDIA Tesla V100 GPU for the   central controller, and  a virtual machine equipped with an NVIDIA GeForce GTX 1070 Ti GPU for the generic source node. Fig.~\ref{fig:robustness}(d) shows the  inference latency for each scheme.

{\bf Lessons learnt on inference latency.}
Note that inference latency increases fairly linearly with the number of deployed source nodes: as the radio challenge becomes more congested the transmission throughput decreases.
The \CentralizedInferenceTag{} and \SelectiveCentralizedInferenceColorHistTag{} schemes perform the worst in terms of inference latency. This is due to the amount of time required to transmit raw views between nodes, which dominates any other latency contribution in our reference scenario. Note that, even if the \SelectiveCentralizedInferenceColorHistTag{} is able to achieve a considerable gain in terms of network overhead, thus reducing bandwidth consumption, it perform worse than \CentralizedInferenceTag{} in terms of latency as the delay due to context sharing adds to the already sizeable time required by the (selected) source nodes to transmit their views.
Inference latency decreases as more computational load is shifted toward the source nodes (either feature extraction of local inference). However, this often comes at the cost of a decrease in prediction accuracy, as in the case of ensemble-based schemes. Both the \EnsembleInferenceTag{} and the \SelectiveCentralizedInferenceEmbeddingTag{} seems to strike a good trade-off between inference latency and accuracy.

\paragraph{\textbf{Robustness to unreliable communication links.}}
\label{par:robustness}
We are also interested in assessing the robustness of the different inference schemes in the presence of unreliable communication links between source nodes and edge servers. The results of the previous experiment can be interpreted considering a reduced number of source nodes $N$ to mimic faults in the communication links between the central controller and a subset of source nodes. Looking at how the accuracy is affected by the partial availability of views,  \SelectiveCentralizedInferenceColorHistTag{} and \SelectiveEnsembleInferenceColorHistTag{}  emerge as the most robust schemes, both boasting values of standard deviation for the accuracy below $2.5\%$ (Table~\ref{tab:exp}).

{\bf Lessons learnt on robustness to link failures.} 
The accuracy of all schemes seems to be affected to some degree by the reduced availability of sources nodes.  However, selective schemes based on color histograms are to be preferred to embedding-based ones when the reliability of the communication links is a concern.
In these schemes in fact, before transmitting their views, all nodes will communicate in order to establish the current context. This context is then used to collectively decide which views should participate in the inference.
Assuming that failures do not occur between this context-sharing phase and the actual transmission of views, any offline node would not contribute to the current context, thus enabling the rest of the nodes to collectively select the more relevant set of views based solely on their present availability.

%\paragraph{\textbf{Robustness to link quality variability and noise.}} All schemes can operate in the presence of low radio link quality, albeit with some degradation in accuracy. Selective schemes based on color histograms prove to be more robust and, thus, more suitable for scenarios with unreliable connectivity between end devices and edge server.  

\begin{figure}
\centering
\resizebox{\columnwidth}{!}{
    \includegraphics[width=\textwidth]{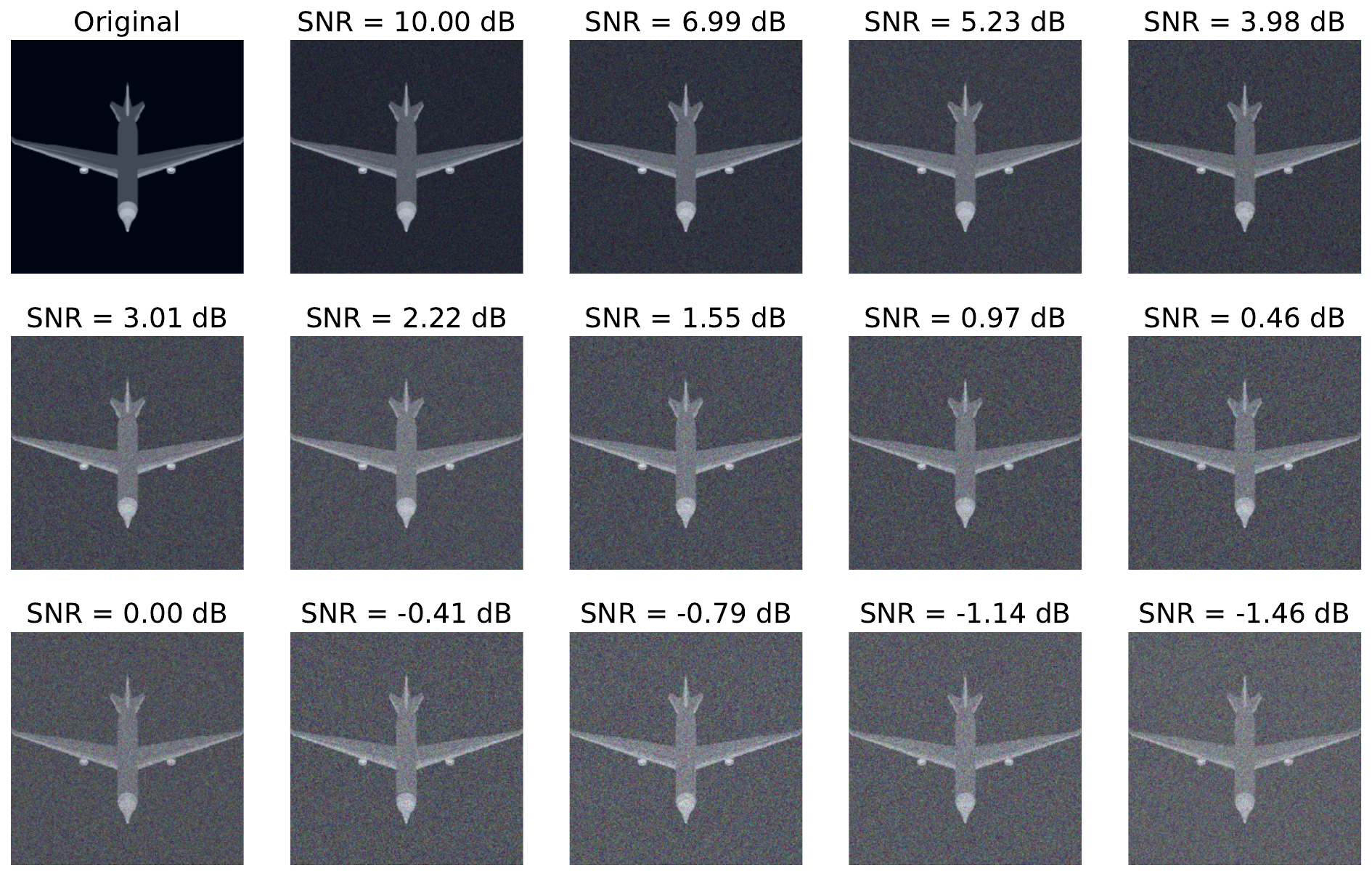}
}
\caption{A view perturbed with additive Gaussian noise attaining different SNR levels.}\label{fig:noise_snr}
\end{figure}

\paragraph{\textbf{Sensitivity to noise.}}
\label{par:sensitivity}
To assess the sensitivity of different inference schemes to noise, we performed another experiment in which we applied Gaussian RGB noise to each sample in the dataset and then tested how the accuracy of the different schemes was affected. Specifically, we applied a random additive noise with zero-mean and different values of standard deviation $\sigma$ to each channel independently. By varying $\sigma$, we generated multiple perturbed datasets with different levels of signal-to-noise ratio (SNR), as depicted in  Fig.~\ref{fig:noise_snr}. 
In Fig.~\ref{fig:noise_curves} we show the accuracy degradation in percentage for each scheme with respect to the case of an unperturbed dataset, as the SNR varies. Each perturbed dataset includes all available views and, as done previously, we use the best-performing similarity threshold for each selective inference scheme. 
The impact of noise on accuracy remains negligible only as long as the SNR level remains between 15 and 20\,dB (i.e., the noise power is within 1\% and 3\% of the power of the original image). A performance degradation starts instead to be evident for lower SNR values (noise power within 3\% and 10\% of the original image power). Then, after decreasing slowly between 10 and 7\,dB of SNR, the accuracy rapidly plummets for SNR below 7\,dB. 
Interestingly,  \CentralizedInferenceTag{}  and  \SelectiveCentralizedInferenceColorHistTag{} prove to be the least sensitive to noise (leftmost curves): they exhibit almost the same sensitivity profile and both retain accuracy levels above 90\% for SNR values up to 11\,dB (i.e., noise power around 8\% of the original image power). Conversely, \SelectiveEnsembleInferenceEmbeddingTag{} and \SelectiveCentralizedInferenceEmbeddingTag{}  show the highest noise sensitivity (rightmost curves). 

{\bf Lessons learnt on noise sensitivity.} 
Regarding noise sensitivity, all schemes prove to be  sensitive to noise in the input data to some degree.
Selective schemes based on color-histograms demonstrate to be substantially more robust than their embedding-based counterparts. This is probably due to how the view descriptors are aggregated to produce the context in the two distinct variants, i.e., through an average in the color histogram case and through a maximum operation in the embedding case, with the latter tending to be more easily affected by noise.
Furthermore, centralized inference schemes seems to be less affected by noise than ensemble ones. This may be due to the fact that, by aggregating the distorted views at a earlier stage, the effect of noise on the features of a view may be mitigated by the features captured by other views before it has a chance to distort the prediction, in the centralized case. 

\begin{figure}
\centering
\resizebox{\columnwidth}{!}{
    \includegraphics[width=\textwidth]{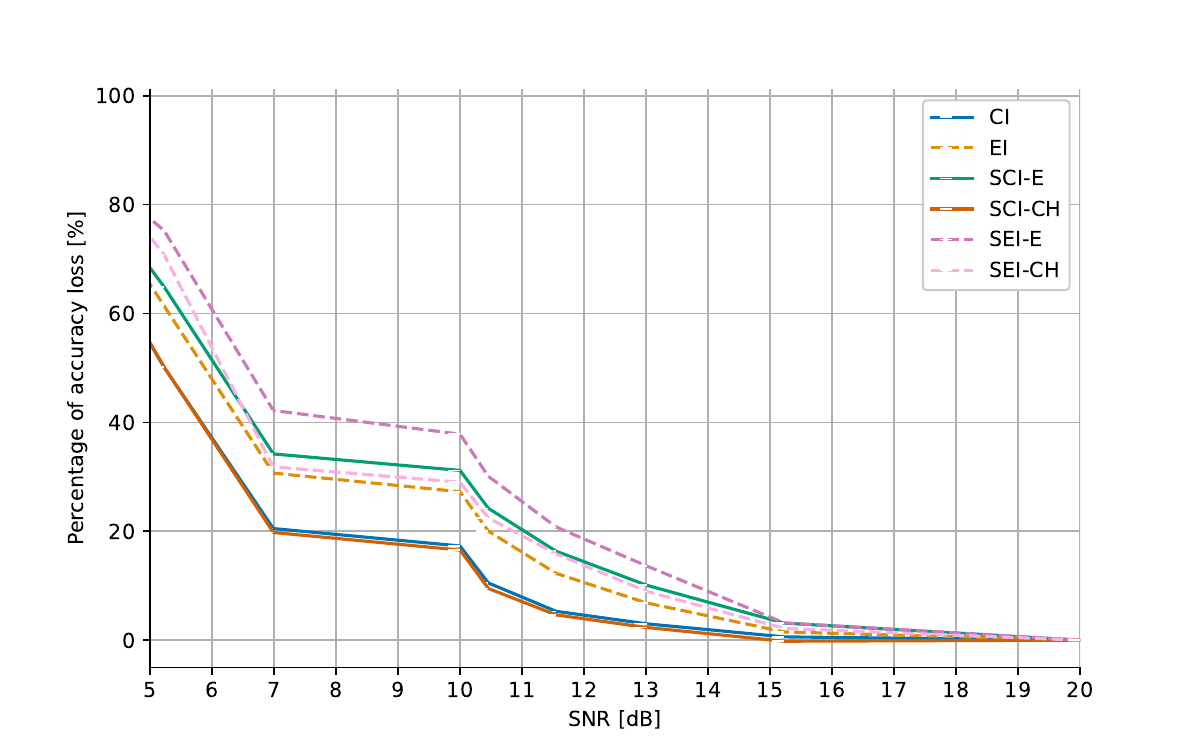}
}
\caption{Comparison among the different schemes in terms of percentage of accuracy loss with respect to the unperturbed case, for varying SNR levels.}\label{fig:noise_curves}
\end{figure}

\begin{table*}
\caption{Qualitative comparison of the considered collaborative inference schemes}
\begin{threeparttable}
\adjustbox{max width=\textwidth}{%
\centering
\SetTblrInner{hspan=minimal}
\begin{tblr}{
  colspec={|c|c|c|c|c|c|c|c|},
  column{4}={2.3cm},
  column{5}={2.1cm},
  width = \linewidth
}
\toprule
\SetCell[r=2]{c} Scheme & \SetCell[r=2]{c} {Prediction~\\accuracy } & \SetCell[r=2]{c} {Communication~\\overhead } & \SetCell[c=2]{c} Computational requirements &        & \SetCell[r=2]{c} {Robustness to~\\link quality} & \SetCell[r=2]{c} {Noise~\\sensitivity} & \SetCell[r=2]{c} {Privacy~\\preservation} \\
\cmidrule{4-5}
& & &  Controller & Source nodes & & & \\
\midrule
CI     & {\textit{High} \\ full MVCNN, \\ use all views} & {\textit{High} \\ nodes share \\ raw images } & {\textit{High} \\ full MVCNN} & {\textit{None} \\ no processing} & {\textit{Medium}} & {\textit{Low}} & {\textit{None}} \\
SCI-E  & {\textit{Medium-High*} \\ split MVCNN, \\ discard views} & {\textit{Medium} \\ nodes share \\ embeddings } & {\textit{Low} \\ view pooling, \\classification} & {\textit{High}: \\ feature extraction} & {\textit{Low}} & {\textit{High}} & {\textit{Partial}}\\
SCI-CH & {\textit{Medium-High*} \\ full MVCNN, \\ discard views} & {\textit{High} \\ nodes share  color \\ hist. and raw images } & {\textit{High} \\ full MVCNN, \\ hist. aggregation} & {\textit{Low}: \\ color histogram} & {\textit{High} }& {\textit{Low}} & {\textit{None}}\\
EI     & {\textit{Medium} \\ multiple SVCNN, \\ use all views} & {\textit{Low} \\ nodes share labels} & {\textit{Low} \\ consensus} & {\textit{High} \\ full SVCNN} & {\textit{Low}} & {\textit{High}} & {\textit{Full}}\\
SEI-E  & {\textit{Low-Medium*} \\ multiple SVCNN, \\ discard views} & {\textit{Medium} \\ nodes share labels \\ and embeddings } & {\textit{Low} \\ consensus, \\ view pooling} & {\textit{High} \\ full SVCNN} & {\textit{Medium}} & {\textit{High}}& {\textit{Partial}}\\
SEI-CH & {\textit{Low-Medium*} \\ multiple SVCNN, \\ discard views} & {\textit{Low} \\ nodes share labels \\ and color hist.} & {\textit{Low} \\ consensus, \\ hist. aggregation} & {\textit{Medium} \\ full SVCNN, \\ color histogram} & {\textit{Low}}& {\textit{High}} & {\textit{Partial}}\\
\bottomrule
\end{tblr}
}%
\begin{tablenotes}\footnotesize
\item[*] Depending on the selected similarity threshold.
\end{tablenotes}
\end{threeparttable}
\label{table:properties}
\end{table*}

{\bf Overall quantitative comparison.} 
At last, we present the qualitative comparison of the considered schemes in Table\,\ref{table:properties}. 
The table  summarizes the  advantages and disadvantages of each schemes, highlighting a set of networking and data-related properties.

% \begin{table}
% \centering
% \caption{{\color{blue}Values on Signal-to-Noise Ratio (SNR), Channel Quality Indicator (CQI) and Modulation Coding Scheme (MCS) index for the UEs acting as source nodes in the reference scenario.}}
% \label{tab:channel_quality}
% {\color{blue}
% \begin{tabular}{cccc}
% \textbf{Source node}    & S\textbf{NR [db]} & \textbf{CQI} & \textbf{MCS Index}  \\
% \hline
% UE1 & 19.2     & 12  & 15         \\
% UE2 & 17.7     & 11  & 13         \\
% UE3 & 7.9      & 6   & 4          \\
% UE4 & 11.6     & 8   & 9          \\
% UE5 & 5.3      & 4   & 2          \\
% UE6 & 10.0     & 7   & 8         
% \end{tabular}
% }
% \end{table}

\section{Discussion and Conclusions}
\label{sec:discussion}

The experimental results presented above clearly indicate that the inference schemes we analysed can achieve different trade-offs between accuracy, communication overhead and latency by leveraging different splits of computation between edge and end devices. In this section we provide an additional qualitative comparison between the aforementioned schemes in terms of computational requirements and privacy preservation. 
%Furthermore, we discuss how the considered schemes can be applied to  practical scenarios, with different node and network capabilities and application requirements. 
Furthermore, we  discuss some relevant applications that can benefit from the proposed schemes (Sec. \ref{sec:applications}) and provide some future research directions (Sec. \ref{sec:directions}). 

\paragraph{\textbf{Computational requirements.}} %The proposed schemes use different splits of computation between the central controller and the source nodes. 
Considering the MVCNN based on VGG-16 used in the experiments, the most computationally intensive task is feature extraction, with the convolutional network accounting for 30.7 GFLOPs per view. In contrast, the view pooling layer and the classification portion of the network account for 0.3 and 239.4 MFLOPs per inference, respectively. The \CentralizedInferenceTag{} scheme clearly minimizes the computational burden allocated to the source nodes, 
%as computation is completely offloaded to the central controller. The 
while \SelectiveCentralizedInferenceColorHistTag{} aims at leveraging the advantages of view selection while keeping the computational requirements at the source nodes low. Conversely, selective, embedding-based schemes  and ensemble schemes require the source nodes to perform feature extraction and are thus suited only for scenarios in which sensing devices are equipped with sufficient  processing capabilities (e.g., connected vehicles). Importantly, the \SelectiveEnsembleInferenceColorHistTag{} scheme mitigates the overall computational burden on source nodes, as only color histogram computation is performed locally when a view gets discarded. 
%To assess which collaborative inference scheme is best suited in which scenario, in this section we provide a qualitative comparison between the aforementioned schemes in terms of  network and data-related properties and requirements. Then we provide some insights on relevant aspects that still need to be addressed in collaborative inference to fully unlock the potential of the proposed schemes.

\paragraph{\textbf{Privacy preservation.}} As the data captured by source nodes may contain private information, some scenarios may be subject to privacy preservation requirements. Clearly, the \CentralizedInferenceTag{} and \SelectiveCentralizedInferenceColorHistTag{} schemes, in which the raw input data is shared between source nodes and the central controller, are intrinsically unable to preserve privacy. On the contrary, the non-selective ensemble inference scheme, in which nodes exchange only predictions, is the most robust to privacy leakage. Schemes in which intermediate representation of views are shared between the nodes, being those embeddings or color-histograms, enhance the level of privacy but are only partially privacy-preserving, as they are not immune to leakage of private information~\cite{dingPrivacyPreservingFeatureExtraction2022}.

%\subsection{Qualitative comparison}
%{\color{purple} I prima paragrafi di questa sottosezione sono un po' ripetitivi (sovrapposti ai commenti fatti in sez. 5). Ma non ho una buona soluzione per questo. Che ne pensate?}
%The schemes proposed in Sect.~\ref{sec:schemes} are each characterized by different properties, making them suited for different types of scenarios.
%Properties are categorized into data properties, driven by the data, and system properties, arising from the system architecture.
%Based on the experimental results and the inherent properties of each scheme we provide an overall qualitative comparisons between them. 

%To provide a meaningful comparison, we focus on the following properties: prediction accuracy, data utilization, communication overhead, computational requirements, privacy preservation, robustness to variability in radio link quality, and noise sensitivity. The results of this comparison are summarized in Table~\ref{table:properties}.

\subsection{Application scenarios}
\label{sec:applications}

Some relevant applications that can greatly benefit from the proposed collaborative inference schemes are discussed below. 
The main strength of centralized inference is the superior prediction accuracy it can provide. The completely centralized \CentralizedInferenceTag{} scheme is thus best  suited for application scenarios characterized by devices with scarce computing capabilities, ample communication bandwidth, and no privacy concerns, such as   Quality Control in smart manufacturing. In this context, indeed,  resource-constrained IoT devices are used to analyze images of food products or drugs packaging to detect defects, inconsistencies, and quality issues.  
Conversely, mission and safety critical applications, like  process automation in smart factories or pedestrian detection in connected autonomous vehicles, are often characterized by ultra-low latency and ultra-high accuracy requirements. In these cases, we advocate the use of the \SelectiveCentralizedInferenceEmbeddingTag{} scheme, in which the availability of contextual information from previous time periods and the feature extraction performed in parallel by source nodes can ensure a lower inference latency, while the use of a conservative similarity threshold will still ensure high prediction accuracy.  Instead, \SelectiveCentralizedInferenceColorHistTag{}, given its low computational requirements on source nodes, low noise sensitivity, is  more suitable for scenarios such as outdoor video-surveillance, in which resource-constrained devices collaborate in a noisy environment.

The main advantages of ensemble inference schemes are their reduced communication overhead, along with the yet good accuracy level and the high degree of privacy they provide. The \EnsembleInferenceTag{} allows for the maximum degree of privacy preservation but at the same time proves to be scarcely robust to link failures and highly sensitive to noise. Thus, it  suits best  applications with strong privacy requirements, reliable connectivity, and low noise such as security surveillance in home automation systems. 
When privacy is an important factor but the computational capability of the devices is limited, the \SelectiveEnsembleInferenceColorHistTag{} scheme is to be preferred. Relevant examples include  in-store retail analytics applications (such as customer footfall analysis, people counting, etc.), or people/vehicle identification in access control systems. 
Finally, urban monitoring applications through connected vehicles or UAVs, for instance to track new stores, road surface conditions, health of urban green spaces, are characterized by both privacy concerns and low connection reliability and, hence, are suitable candidates for the \SelectiveEnsembleInferenceEmbeddingTag{} scheme.

\subsection{Challenges and future directions}
\label{sec:directions}

We advocate that collaborative inference schemes, in which nodes share semantically correlated data and computation to make a prediction, can allow an increasingly larger set of fine-grained inference tasks to be pushed from the cloud domain to the network edge. However, a number of challenges should be addressed in order to make these schemes easier to apply in real-world scenarios. 

\paragraph{\textbf{Node selection.}} We considered a scenario in which a set of end devices capture data that are spatially correlated. This easily holds in contexts with static end nodes whose cameras are deployed at fixed locations and orientations, in which case the spatial correlation between sensed data can be determined a priori. In reality, we often deal with hybrid systems comprising both static and mobile nodes, in which the correlation between the sensed data often  arises in real time and most likely involves only a subset of the system nodes at a time. In these cases, how to select  the nodes that should contribute to a collaborative multi-view classification task is still an open issue.

\paragraph{\textbf{Adapt to dynamic network conditions.}} As detailed previously, different collaborative inference schemes, with different parameters, may lead to more or less favorable trade-offs between accuracy,  communication overhead and latency, depending on the condition of the network at a given time. However, especially in a context involving mobile nodes, the underlying network conditions may vary continuously, making the choice of which scheme (or scheme's parametrization) to adopt quite hard. Future research should thus envision methods to allow inference schemes to adapt dynamically to the varying conditions of the network. In particular, when a selective inference scheme is used,  its similarity threshold should be dynamically modulated based on the available bandwidth. 

\paragraph{\textbf{Computational splits and temporal correlation.}}
In our study, we focused on the most relevant   splits of computation between nodes. For instance, in the \SelectiveCentralizedInferenceEmbeddingTag{} scheme, each node executes a multi-view CNN's full feature extraction pipeline. However, more elaborated splits are possible depending on the processing capabilities of the nodes, e.g., executing only a subset of the first convolutional blocks at the source nodes while delegating the remaining feature extraction to the central controller. Additionally, state-of-the-art DNN compression techniques such as model pruning~\cite{wenLearningStructuredSparsity2016} and knowledge distillation~\cite{gouKnowledgeDistillationSurvey2021} may further broaden the scope of applicability of inference schemes by reducing their requirements in terms of processing capabilities. In our current setting, deciding which scheme to use, how the computation should be split between the system nodes, and whether to use data and/or DNN compression techniques would all be one-time decisions made at the system design time. Thus, it would be important to define new methods to let the system make these decisions  automatically, based on the availability of resources and the presence of overarching system requirements, both at a preliminary "negotiation" phase 
and (potentially) at a later stage during the network operation.
Finally, in our study we mainly focused on exploiting spatial correlation between views. An additional aspect to consider   would be to also integrate techniques like frame differentiation~\cite{chenGLIMPSEContinuousRealTime2016}, taking into account temporal correlation as well, in order to further reduce redundant information.

%In particular  higher-level algorithm framework able to automatically select the most suitable scheme and allowing for a more flexible partition of computation between nodes based on their availability of resources and overarching system requirements. 

%Mission critical applications: Smart manufacturing (e.g., machinery control) and Autonomous driving connected vehicles: SCI-E (low latency, high accuracy)

%Videosuirvelliance: SCI-CH (if IoT with lower computational capability and outdoor location because it is robust to noise) OR EI (when low bandwidth availability and/or privacy is needed) 

%Quality Control in which AI analyze images of food products or drugs packaging to detect defects, inconsistencies, and quality issues SEI-CH (IoTs with lower computational capability and low bandwidth consumption)

%urban monitoring through connected vehicles or UAVs (images of new stores, road issues, (with UAVs) vehicular traffic monitoring): SEI-E (low accuracy but medium robustness to link quality variability and partial privacy)

%\section{Conclusions}
%\label{sec:conclusions}
%As of now, the result of quality estimation of views is used only for discriminating whether or not each source node should send its locally-captured view to the central controller. In future work we plan to address the reuse of quality estimation results to aid the classification task.

\section{Acknowledgements}
\label{sec:ack}
This work was supported by the European Commission under Grant Agreement No.\,101095363 (ADROIT6G project) and Grant Agreement No.\,101139266 (6G-INTENSE project).

\bibliographystyle{elsarticle-num} 
% \bibliography{biblio}

\begin{thebibliography}{10}
  \expandafter\ifx\csname url\endcsname\relax
    \def\url#1{\texttt{#1}}\fi
  \expandafter\ifx\csname urlprefix\endcsname\relax\def\urlprefix{URL }\fi
  \expandafter\ifx\csname href\endcsname\relax
    \def\href#1#2{#2} \def\path#1{#1}\fi
  
  \bibitem{malandrinoMatchingDNN2022}
  F.~Malandrino, G.~D. Giacomo, A.~Karamzade, M.~Levorato, C.-F. Chiasserini,
    \href{https://api.semanticscholar.org/CorpusID:254247266}{Matching dnn
    compression and cooperative training with resources and data availability},
    IEEE INFOCOM 2023 - IEEE Conference on Computer Communications (2022) 1--10.
  \newline\urlprefix\url{https://api.semanticscholar.org/CorpusID:254247266}
  
  \bibitem{chenDeepLearningEdge2019}
  J.~Chen, X.~Ran, Deep {{Learning With Edge Computing}}: {{A Review}},
    Proceedings of the IEEE 107~(8) (2019) 1655--1674.
  \newblock \href {https://doi.org/10.1109/JPROC.2019.2921977}
    {\path{doi:10.1109/JPROC.2019.2921977}}.
  
  \bibitem{pulighedduSemanticORAN2023}
  C.~Puligheddu, J.~Ashdown, C.~F. Chiasserini, F.~Restuccia, Sem-o-ran: Semantic
    o-ran slicing for mobile edge offloading of computer vision tasks, IEEE
    Transactions on Mobile Computing (2023) 1--16\href
    {https://doi.org/10.1109/TMC.2023.3339056}
    {\path{doi:10.1109/TMC.2023.3339056}}.
  
  \bibitem{sunSurveyMultiviewMachine2013}
  S.~Sun, A {{Survey}} of {{Multi-view Machine Learning}}, Neural Computing and
    Applications 23~(7-8) (2013) 2031--2038.
  \newblock \href {https://doi.org/10.1007/s00521-013-1362-6}
    {\path{doi:10.1007/s00521-013-1362-6}}.
  
  \bibitem{seelandMultiviewClassificationConvolutional2021}
  M.~Seeland, P.~M{\"a}der, Multi-view {{Classification}} with {{Convolutional
    Neural Networks}}, PLOS ONE 16~(1) (2021) e0245230.
  \newblock \href {https://doi.org/10.1371/journal.pone.0245230}
    {\path{doi:10.1371/journal.pone.0245230}}.
  
  \bibitem{suMultiviewConvolutionalNeural2015}
  H.~Su, S.~Maji, E.~Kalogerakis, E.~{Learned-Miller}, Multi-view {{Convolutional
    Neural Networks}} for {{3D Shape Recognition}}, in: 2015 {{IEEE International
    Conference}} on {{Computer Vision}} ({{ICCV}}), {IEEE}, {Santiago, Chile},
    2015, pp. 945--953.
  \newblock \href {https://doi.org/10.1109/ICCV.2015.114}
    {\path{doi:10.1109/ICCV.2015.114}}.
  
  \bibitem{renSurveyCollaborativeDNN2023}
  W.-Q. Ren, Y.-B. Qu, C.~Dong, Y.-Q. Jing, H.~Sun, Q.-H. Wu, S.~Guo, A
    {{Survey}} on {{Collaborative DNN Inference}} for {{Edge Intelligence}},
    Machine Intelligence Research 20~(3) (2023) 370--395.
  \newblock \href {https://doi.org/10.1007/s11633-022-1391-7}
    {\path{doi:10.1007/s11633-022-1391-7}}.
  
  \bibitem{zhou2019}
  Z.~Zhou, X.~Chen, E.~Li, L.~Zeng, K.~Luo, J.~Zhang, Edge intelligence: Paving
    the last mile of artificial intelligence with edge computing, Proceedings of
    the IEEE 107~(8) (2019) 1738--1762.
  \newblock \href {https://doi.org/10.1109/JPROC.2019.2918951}
    {\path{doi:10.1109/JPROC.2019.2918951}}.
  
  \bibitem{eshratifar2018}
  A.~E. Eshratifar, M.~Pedram,
    \href{https://doi.org/10.1145/3194554.3194565}{Energy and performance
    efficient computation offloading for deep neural networks in a mobile cloud
    computing environment}, in: Proceedings of the 2018 Great Lakes Symposium on
    VLSI, GLSVLSI '18, Association for Computing Machinery, New York, NY, USA,
    2018, p. 111–116.
  \newblock \href {https://doi.org/10.1145/3194554.3194565}
    {\path{doi:10.1145/3194554.3194565}}.
  \newline\urlprefix\url{https://doi.org/10.1145/3194554.3194565}
  
  \bibitem{kang2017}
  Y.~Kang, J.~Hauswald, C.~Gao, A.~Rovinski, T.~Mudge, J.~Mars, L.~Tang,
    \href{https://doi.org/10.1145/3093337.3037698}{Neurosurgeon: Collaborative
    intelligence between the cloud and mobile edge}, SIGARCH Comput. Archit. News
    45~(1) (2017) 615–629.
  \newblock \href {https://doi.org/10.1145/3093337.3037698}
    {\path{doi:10.1145/3093337.3037698}}.
  \newline\urlprefix\url{https://doi.org/10.1145/3093337.3037698}
  
  \bibitem{wangRFSensing2018}
  X.~Wang, X.~Wang, S.~Mao, Rf sensing in the internet of things: A general deep
    learning framework, IEEE Communications Magazine 56~(9) (2018) 62--67.
  \newblock \href {https://doi.org/10.1109/MCOM.2018.1701277}
    {\path{doi:10.1109/MCOM.2018.1701277}}.
  
  \bibitem{malkaDecentralized2022}
  M.~Malka, E.~Farhan, H.~Morgenstern, N.~Shlezinger, Decentralized low-latency
    collaborative inference via ensembles on the edge (2022).
  \newblock \href {http://arxiv.org/abs/2206.03165} {\path{arXiv:2206.03165}}.
  
  \bibitem{zeng2020}
  L.~Zeng, X.~Chen, Z.~Zhou, L.~Yang, J.~Zhang, Coedge: Cooperative dnn inference
    with adaptive workload partitioning over heterogeneous edge devices, IEEE/ACM
    Transactions on Networking 29~(2) (2021) 595--608.
  \newblock \href {https://doi.org/10.1109/TNET.2020.3042320}
    {\path{doi:10.1109/TNET.2020.3042320}}.
  
  \bibitem{maoSurveyOnMobile2017}
  Y.~Mao, C.~You, J.~Zhang, K.~Huang, K.~B. Letaief, A survey on mobile edge
    computing: The communication perspective, IEEE Communications Surveys \&
    Tutorials 19~(4) (2017) 2322--2358.
  \newblock \href {https://doi.org/10.1109/COMST.2017.2745201}
    {\path{doi:10.1109/COMST.2017.2745201}}.
  
  \bibitem{cohenLightweightCompression2021}
  R.~Cohen, H.~Choi, I.~Bajic, Lightweight compression of intermediate neural
    network features for collaborative intelligence, IEEE Open Journal of
    Circuits and Systems 2 (2021) 350--362.
  \newblock \href {https://doi.org/10.1109/OJCAS.2021.3072884}
    {\path{doi:10.1109/OJCAS.2021.3072884}}.
  
  \bibitem{choudharyComprehensiveSurveyModelCompression2020}
  T.~Choudhary, V.~K. Mishra, A.~Goswami, S.~Jagannathan,
    \href{https://api.semanticscholar.org/CorpusID:211062209}{A comprehensive
    survey on model compression and acceleration}, Artificial Intelligence Review
    (2020) 1--43.
  \newline\urlprefix\url{https://api.semanticscholar.org/CorpusID:211062209}
  
  \bibitem{hao2023}
  Z.~Hao, G.~Xu, Y.~Luo, H.~Hu, J.~An, S.~Mao,
    \href{https://doi.org/10.1109/TMC.2022.3183098}{Multi-agent collaborative
    inference via dnn decoupling: Intermediate feature compression and edge
    learning}, IEEE Transactions on Mobile Computing 22~(10) (2023) 6041–6055.
  \newblock \href {https://doi.org/10.1109/TMC.2022.3183098}
    {\path{doi:10.1109/TMC.2022.3183098}}.
  \newline\urlprefix\url{https://doi.org/10.1109/TMC.2022.3183098}
  
  \bibitem{wangMultiAgentSystemsCollaborative2024}
  S.~Wang, Y.~Jing, K.~Wang, X.~Wang, Multi-{{Agent Systems}} for {{Collaborative
    Inference Based}} on {{Deep Policy Q-Inference Network}}, Journal of Grid
    Computing 22~(1) (2024) 38.
  \newblock \href {https://doi.org/10.1007/s10723-024-09750-w}
    {\path{doi:10.1007/s10723-024-09750-w}}.
  
  \bibitem{yaoDeepSense2017}
  S.~Yao, S.~Hu, Y.~Zhao, A.~Zhang, T.~Abdelzaher,
    \href{https://doi.org/10.1145/3038912.3052577}{Deepsense: A unified deep
    learning framework for time-series mobile sensing data processing}, in:
    Proceedings of the 26th International Conference on World Wide Web, WWW '17,
    International World Wide Web Conferences Steering Committee, Republic and
    Canton of Geneva, CHE, 2017, p. 351–360.
  \newblock \href {https://doi.org/10.1145/3038912.3052577}
    {\path{doi:10.1145/3038912.3052577}}.
  \newline\urlprefix\url{https://doi.org/10.1145/3038912.3052577}
  
  \bibitem{qiuKestrelVideoAnalytics2018}
  H.~Qiu, X.~Liu, S.~Rallapalli, A.~J. Bency, K.~Chan, R.~Urgaonkar,
    B.~Manjunath, R.~Govindan, Kestrel: {{Video Analytics}} for {{Augmented
    Multi-Camera Vehicle Tracking}}, in: 2018 {{IEEE}}/{{ACM Third International
    Conference}} on {{Internet-of-Things Design}} and {{Implementation}}
    ({{IoTDI}}), {IEEE}, {Orlando, FL}, 2018, pp. 48--59.
  \newblock \href {https://doi.org/10.1109/IoTDI.2018.00015}
    {\path{doi:10.1109/IoTDI.2018.00015}}.
  
  \bibitem{yanDeepMultiviewLearning2021}
  X.~Yan, S.~Hu, Y.~Mao, Y.~Ye, H.~Yu, Deep {{Multi-view Learning Methods}}: {{A
    Review}}, Neurocomputing 448 (2021) 106--129.
  \newblock \href {https://doi.org/10.1016/j.neucom.2021.03.090}
    {\path{doi:10.1016/j.neucom.2021.03.090}}.
  
  \bibitem{silvaMultiViewFineGrained2021}
  B.~Silva, F.~R. Barbosa-Anda, J.~Batista, Multi-view fine-grained vehicle
    classification with multi-loss learning, in: 2021 IEEE International
    Conference on Autonomous Robot Systems and Competitions (ICARSC), 2021, pp.
    209--214.
  \newblock \href {https://doi.org/10.1109/ICARSC52212.2021.9429780}
    {\path{doi:10.1109/ICARSC52212.2021.9429780}}.
  
  \bibitem{chenGLIMPSEContinuousRealTime2016}
  T.~Y.-H. Chen, H.~Balakrishnan, L.~Ravindranath, P.~Bahl, {{GLIMPSE}}:
    {{Continuous}}, {{Real-Time Object Recognition}} on {{Mobile Devices}},
    GetMobile: Mobile Comp. and Comm. 20~(1) (2016) 26--29.
  \newblock \href {https://doi.org/10.1145/2972413.2972423}
    {\path{doi:10.1145/2972413.2972423}}.
  
  \bibitem{lecun2015deep}
  Y.~LeCun, Y.~Bengio, G.~Hinton, Deep learning, nature 521~(7553) (2015) 436.
  
  \bibitem{krizhevskyImagenetClassification2012}
  A.~Krizhevsky, I.~Sutskever, G.~E. Hinton,
    \href{https://proceedings.neurips.cc/paper_files/paper/2012/file/c399862d3b9d6b76c8436e924a68c45b-Paper.pdf}{Imagenet
    classification with deep convolutional neural networks}, in: F.~Pereira,
    C.~Burges, L.~Bottou, K.~Weinberger (Eds.), Advances in Neural Information
    Processing Systems, Vol.~25, Curran Associates, Inc., 2012.
  \newline\urlprefix\url{https://proceedings.neurips.cc/paper_files/paper/2012/file/c399862d3b9d6b76c8436e924a68c45b-Paper.pdf}
  
  \bibitem{standard2007colorimetry}
  C.~Standard, et~al., Colorimetry-part 4: Cie 1976 l* a* b* colour space,
    International Standard (2007) 2019--06.
  
  \bibitem{yangLargeScaleCarDataset2015}
  L.~Yang, P.~Luo, C.~C. Loy, X.~Tang,
    \href{http://dblp.uni-trier.de/db/conf/cvpr/cvpr2015.html#YangLLT15}{A
    large-scale car dataset for fine-grained categorization and verification.},
    in: CVPR, IEEE Computer Society, 2015, pp. 3973--3981.
  \newline\urlprefix\url{http://dblp.uni-trier.de/db/conf/cvpr/cvpr2015.html#YangLLT15}
  
  \bibitem{manjunathColorTextureDescriptors2001}
  B.~Manjunath, J.-R. Ohm, V.~Vasudevan, A.~Yamada, Color and texture
    descriptors, IEEE Transactions on Circuits and Systems for Video Technology
    11~(6) (2001) 703--715.
  \newblock \href {https://doi.org/10.1109/76.927424}
    {\path{doi:10.1109/76.927424}}.
  
  \bibitem{swainColorIndexing1991}
  M.~J. Swain, D.~H. Ballard, Color indexing, International Journal of Computer
    Vision 7~(1) (1991) 11--32.
  \newblock \href {https://doi.org/10.1007/BF00130487}
    {\path{doi:10.1007/BF00130487}}.
  
  \bibitem{zhirong3DShapeNets2015}
  Z.~Wu, S.~Song, A.~Khosla, F.~Yu, L.~Zhang, X.~Tang, J.~Xiao, 3d shapenets: A
    deep representation for volumetric shapes, in: 2015 IEEE Conference on
    Computer Vision and Pattern Recognition (CVPR), 2015, pp. 1912--1920.
  \newblock \href {https://doi.org/10.1109/CVPR.2015.7298801}
    {\path{doi:10.1109/CVPR.2015.7298801}}.
  
  \bibitem{suDeeperLookAt3DShape2018}
  J.-C. Su, M.~Gadelha, R.~Wang, S.~Maji, A deeper look at 3d shape classifiers,
    in: Proceedings of the European Conference on Computer Vision (ECCV)
    Workshops, 2018.
  
  \bibitem{simonyanVeryDeepConvolutional2015}
  K.~Simonyan, A.~Zisserman, \href{http://arxiv.org/abs/1409.1556}{Very deep
    convolutional networks for large-scale image recognition}, in: Y.~Bengio,
    Y.~LeCun (Eds.), 3rd International Conference on Learning Representations,
    {ICLR} 2015, San Diego, CA, USA, May 7-9, 2015, Conference Track Proceedings,
    2015.
  \newline\urlprefix\url{http://arxiv.org/abs/1409.1556}
  
  \bibitem{dengImagenet2009}
  J.~Deng, W.~Dong, R.~Socher, L.-J. Li, K.~Li, L.~Fei-Fei, Imagenet: A
    large-scale hierarchical image database, in: 2009 IEEE conference on computer
    vision and pattern recognition, Ieee, 2009, pp. 248--255.
  
  \bibitem{dingPrivacyPreservingFeatureExtraction2022}
  X.~Ding, H.~Fang, Z.~Zhang, K.~Choo, H.~Jin, Privacy-preserving feature
    extraction via adversarial training, IEEE Transactions on Knowledge \& Data
    Engineering 34~(04) (2022) 1967--1979.
  \newblock \href {https://doi.org/10.1109/TKDE.2020.2997604}
    {\path{doi:10.1109/TKDE.2020.2997604}}.
  
  \bibitem{wenLearningStructuredSparsity2016}
  W.~Wen, C.~Wu, Y.~Wang, Y.~Chen, H.~Li, Learning structured sparsity in deep
    neural networks, in: Proceedings of the 30th International Conference on
    Neural Information Processing Systems, NIPS'16, Curran Associates Inc., Red
    Hook, NY, USA, 2016, p. 2082–2090.
  
  \bibitem{gouKnowledgeDistillationSurvey2021}
  J.~Gou, B.~Yu, S.~J. Maybank, D.~Tao,
    \href{https://doi.org/10.1007/s11263-021-01453-z}{Knowledge distillation: A
    survey}, Int. J. Comput. Vision 129~(6) (2021) 1789–1819.
  \newblock \href {https://doi.org/10.1007/s11263-021-01453-z}
    {\path{doi:10.1007/s11263-021-01453-z}}.
  \newline\urlprefix\url{https://doi.org/10.1007/s11263-021-01453-z}
  
  \end{thebibliography}

\end{document}